\documentclass[10pt,letterpaper]{article}
\usepackage[top=0.85in,left=2.75in,footskip=0.75in]{geometry}

\usepackage{amsmath,amssymb}

\usepackage{changepage}

\usepackage[utf8x]{inputenc}

\usepackage{textcomp,marvosym}

\usepackage{cite}

\usepackage{nameref,hyperref}

\usepackage[right]{lineno}

\usepackage{microtype}
\DisableLigatures[f]{encoding = *, family = * }

\usepackage[table]{xcolor}

\usepackage{array}

\newcolumntype{+}{!{\vrule width 2pt}}

\newlength\savedwidth

\raggedright
\usepackage[aboveskip=1pt,labelfont=bf,labelsep=period,justification=raggedright,singlelinecheck=off]{caption}

\makeatletter
\renewcommand{\@biblabel}[1]{\quad#1.}
\makeatother

\usepackage{lastpage,fancyhdr,graphicx}
\usepackage{epstopdf}
\fancyheadoffset[L]{2.25in}
\fancyfootoffset[L]{2.25in}
\begin{document}
\vspace*{0.2in}

\begin{flushleft}
{\Large
\textbf\newline{Modeling human intuitions about liquid flow with particle-based simulation} 
}
\newline
\\
Christopher J. Bates \textsuperscript{1\Yinyang\textcurrency a *},
Ilker Yildirim\textsuperscript{1},
Joshua B. Tenenbaum\textsuperscript{1},
Peter Battaglia\textsuperscript{1\textcurrency b \Yinyang},
\\
\bigskip
\textbf{1} Department of Brain and Cognitive Sciences, MIT, Cambridge, Massachusetts, USA
\\
\bigskip

%
%
\Yinyang These authors contributed equally to this work.


\textcurrency a Current Address: Department of Brain and Cognitive Sciences, University of Rochester, Rochester, New York

\textcurrency b Current Address: Google DeepMind, London, United Kingdom 



* cjbates@ur.rochester.edu

\end{flushleft}
\section*{Abstract}
\noindent Humans can easily describe, imagine, and, crucially, \textit{predict} a wide variety of behaviors of liquids--splashing, squirting, gushing, sloshing, soaking, dripping, draining, trickling, pooling, and pouring--despite tremendous variability in their material and dynamical properties. Here we propose and test a computational model of how people perceive and predict these liquid dynamics, based on coarse approximate simulations of fluids as collections of interacting particles. Our model is analogous to a ``game engine in the head'', drawing on techniques for interactive simulations (as in video games) that  optimize for efficiency and natural appearance rather than physical accuracy.  In two behavioral experiments, we found that the model accurately captured people's predictions about how liquids flow among complex solid obstacles, and was significantly better than two alternatives based on simple heuristics and deep neural networks. Our model was also able to explain how people's predictions varied as a function of the liquids' properties (e.g., viscosity and stickiness). Together, the model and empirical results extend the recent proposal that human physical scene understanding for the dynamics of rigid, solid objects can be supported by approximate probabilistic simulation, to the more complex and unexplored domain of fluid dynamics.

\section*{Author summary}
Although most people struggle to learn physics in school, every human brain is a remarkable “intuitive physicist” when it comes to the quick, unconscious judgments we make in interacting with the world. Without effort, and with surprisingly high quantitative accuracy, we can judge when a plate placed near the edge of a table might be at risk of falling, or how far a glass filled with a certain amount of water can be tipped before the water is in danger of spilling.  What kinds of computations in the brain support these abilities? We suggest an answer based on probabilistic inference operating over particle-based simulations, the same class of approximation methods used in video games to simulate convincing real-time interactions between objects in a virtual environment.  This hypothesis can potentially account for people's quantitative, graded judgments in diverse and novel situations including a wide array of materials and physical properties, without positing a large number of separate systems or heuristics. Here, we build on previous evidence that a system of approximate probabilistic simulation supports judgments about rigid objects (e.g. judging the stability of towers of blocks, as in the game Jenga), and ask whether people can also make systematic and accurate predictions about flowing and splashing liquids, such as water or honey.  We show that it is possible to capture people's quantitative predictions using a computational model that approximates the true underlying fluid dynamics to varying degrees of coarseness, and find that people’s responses are most consistent with a very coarse approximation; while typical engineering applications might use tens or hundreds of thousands of particles to simulate a fluid, the brain might get by with roughly a hundred particles.  Furthermore, we find that people consistently underestimate the potential energy of a splashing liquid in our virtual scenes, and that our model  captures this behavior.


\section*{Introduction}
\begin{figure}[h]
\centering
\includegraphics[width=0.5\columnwidth]{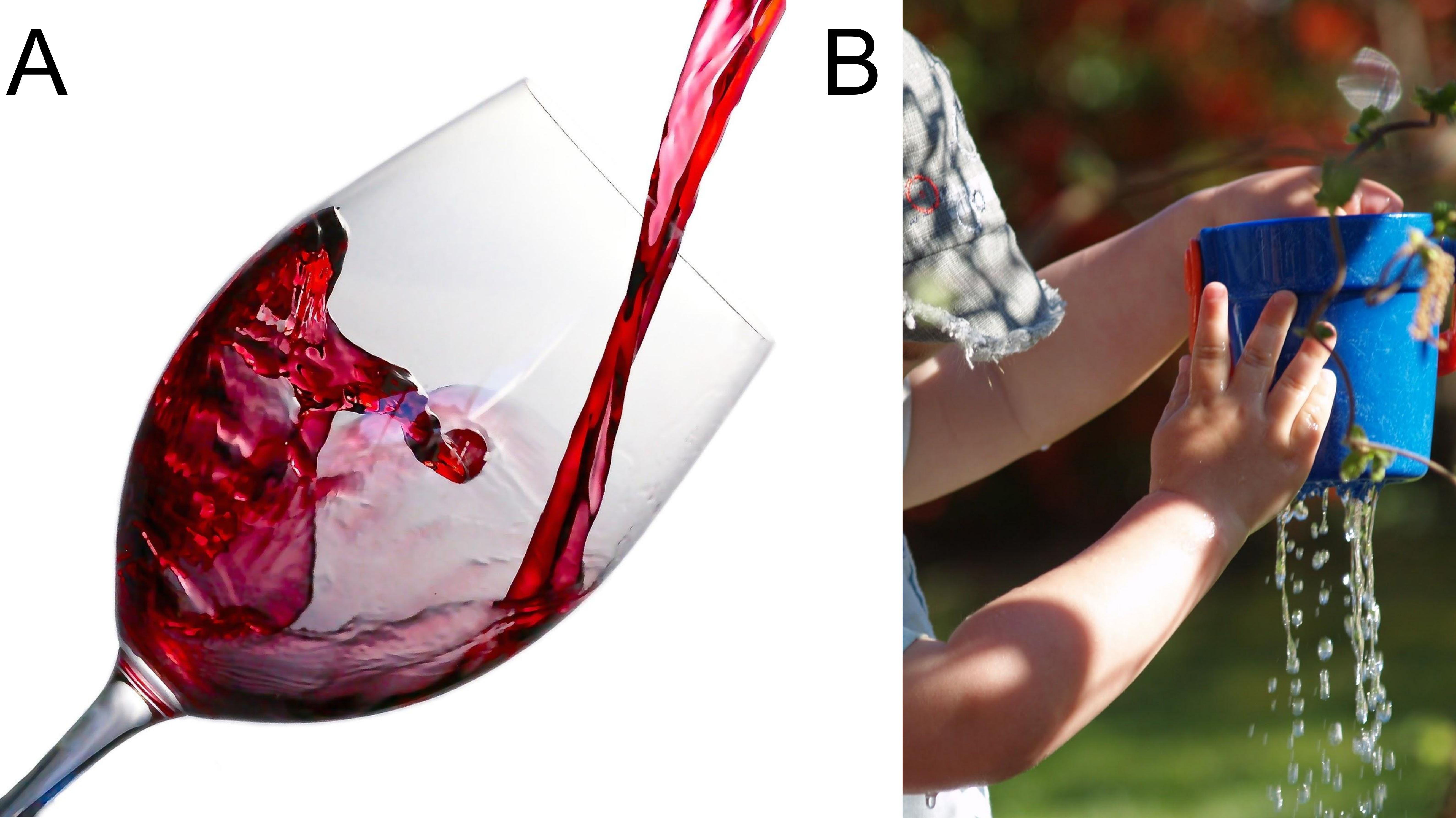}
\caption{\label{fig:intro} \small (\textit{A}) Dynamic fluids are very complex, yet ubiquitous in everyday scenes. (\textit{B}) Humans--even young children--can reason about and interact with liquids effectively.}
\end{figure}

From a glance at liquid flowing into a glass (Fig~\ref{fig:intro}A), you can infer a great deal: it is pouring rapidly, likely from a spout or small opening; it is not viscous; little will likely splash out, although if the angle of the glass were lowered slightly, perhaps much more might escape.  Even young children can perceive and interact with liquids in motion  (Fig~\ref{fig:intro}B) in ways well beyond the capabilities of modern robots and artificial intelligence (AI) systems.  How do people draw rich intuitions about liquids? What is the nature of people's implicit knowledge of liquid dynamics? By what mechanisms is this knowledge applied to support their everyday interactions with liquids? 

A growing body of evidence supports the view that humans have rich knowledge of everyday rigid body physics, which inform their predictions, inferences, and planning through a system of probabilistic inference \cite{sanborn2013reconciling, sanborn2014testing, battaglia2013simulation, gerstenberg2012noisy, smith2013consistent, smith2013sources, hamrick2016inferring}. \cite{battaglia2013simulation} proposed a cognitive mechanism for physical scene understanding based on  ``approximate probabilistic simulation''. They posited that objects' spatial geometry and physical attributes, as well as certain laws of mechanics, are represented approximately, and support fast, efficient probabilistic judgments over short time scales via small numbers of simulations based on sampled estimates of the underlying world state. Their model of human cognition's ``intuitive physics engine'' explained people's physical predictions about stability and support relationships, and the motion of objects under gravity, across a wide range of rigid body scenes.

\begin{figure}[ht!]
\centering
\includegraphics[width=1\linewidth]{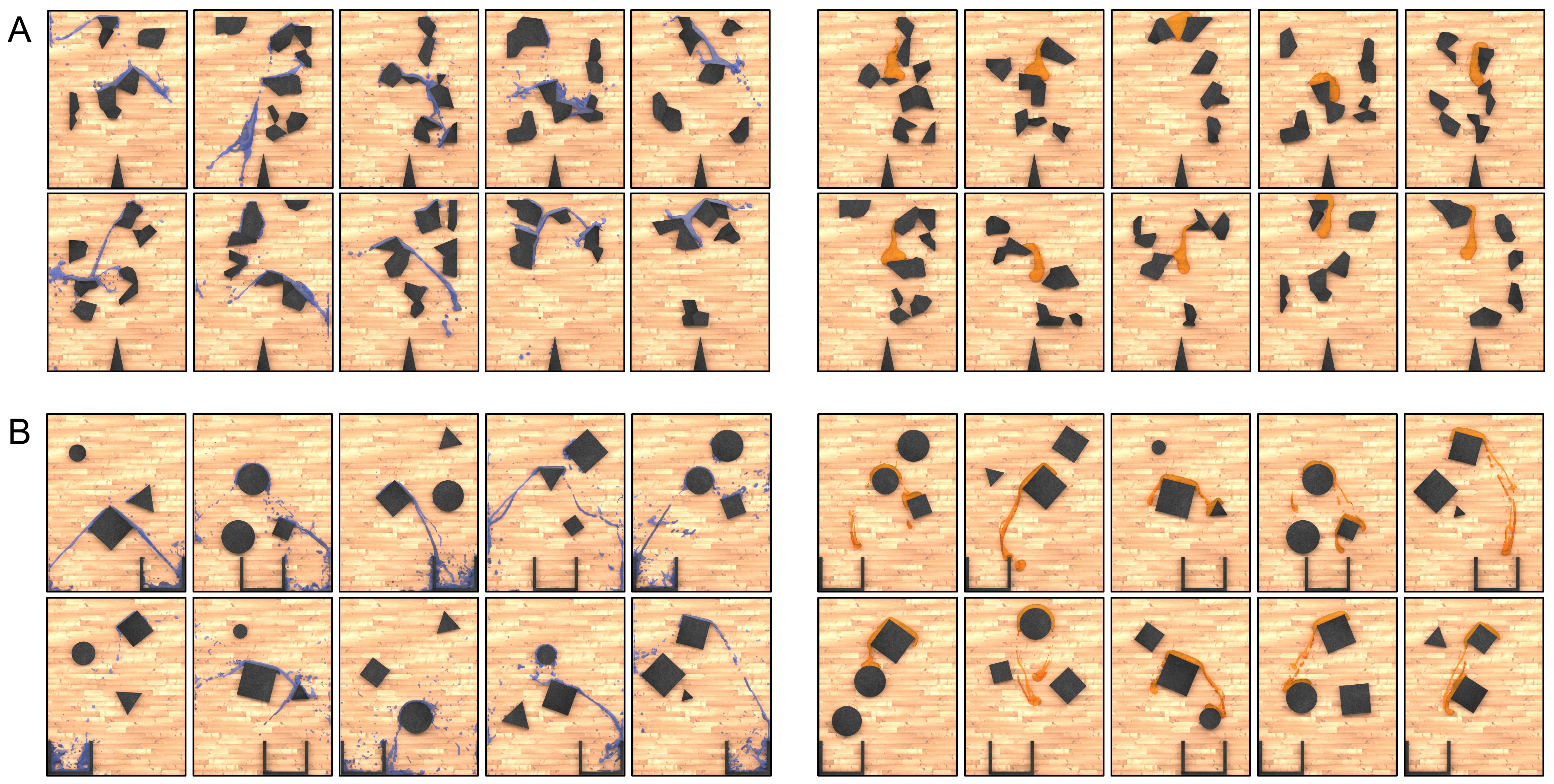}
\caption{\label{fig:stimuli} \small Experimental stimuli. (\textit{A}) Examples of water and honey stimuli from Experiment 1 after gravity was applied. (\textit{B}) Examples of water and honey stimuli from Experiment 2. (Subjects only saw liquid in motion as part of the practice phase, and otherwise only saw a static image of the liquid in its starting position.)}
\end{figure}

Here we explore how people understand physical scenes involving fluids moving around rigid objects, modeling their judgments as being driven by, in part, some form of approximate probabilistic simulations. We asked three related questions: How well can people predict the behavior of liquids in complex scenes? How well can approximate simulation-based models account for people's quantitative predictions across a range of scenes and different kinds of liquids? How do people's uncertainty and computational resource limitations influence their judgments? We tackled these questions by probing people's predictive judgments in two psychophysical experiments, and comparing human predictions to five different models, including four different kinds of simulation-based models varying in their computational complexity and physical assumptions, 
and a simple neural network alternative that approaches the problem as a pattern recognition task, using a deep convolutional network of the kind that has recently been successful in many computer vision applications. We examined how well people's predictions matched each model, and how they co-varied with each model as a function of liquid properties, such as viscosity and stickiness. 

Our experiments asked human participants to predict how a liquid would flow through and around complex arrangements of obstacles (Figure~\ref{fig:stimuli}).  We used two related tasks representative of different kinds of real-world judgments. In the first experiment (example stimuli shown in Figure~\ref{fig:stimuli}A), participants were asked to predict what percentage of the liquid would end up on the right side of the divider. In the second experiment (example stimuli shown in Figure~\ref{fig:stimuli}B), participants were asked what percentage would end up in the cup.
In each experiment, we used computer-animated liquid stimuli with both water-like and honey-like properties.  This allowed us to test both people's abilities to predict flow for a range of different everyday liquids, and different models' abilities to capture human intuitions across that range.

We compared people's judgments to a spectrum of simulation-based computational models, as well as a neural network alternative. Our two most sophisticated models are particle simulation models inspired by approximations to physics-based models used in computer graphics, video games, and computational fluid dynamics: the ``Intuitive Fluids Engine'' (IFE) model, based on smooth particle hydrodynamics (Figure~\ref{fig:model}), is the most physically appropriate for our tasks, although it is also the most computationally complex.  The other physics-based simulation account,  ``MarbleSim'', is similar but uses rigid body interactions between particles.  We also consider a heuristic dynamic simulation model (``SimpleSim'') that is computationally simpler than either physics-based model, and an even simpler heuristic simulator (the ``Gravity heuristic'') that is not based on a dynamic simulation at all, but only on the geometric principle that particles under gravity move downwards unless blocked by obstacles.  
All the dynamics-based models had degrees of freedom permitting them to make more water-like or honey-like predictions, and we explored to what extent varying these parameters could let these models capture people's varying intuitions for these different kinds of liquids. 

To preview our results, we found that the more physically motivated particle-based simulation models provided the best single account of people's predictions across all our experimental conditions.  These models offer a general-purpose computational account of how people predict the motion of fluids on short time scales, in a similar spirit to \cite{battaglia2013simulation}'s approximate probabilistic simulation models of intuitive physics for solid rigid bodies. The flowing liquids in our experimental scenes are much more complex, with far more degrees of freedom, than the rigid body systems on which \cite{battaglia2013simulation}'s original ``intuitive physics engine'' work focused, but our ``intuitive fluids engine'' (IFE) model is developed in a similar spirit.  The model is based on a commonly used method for simulating fluids, based on collections of dynamic, interacting particles, which is theoretically appealing on several grounds: It can capture a wide variety of materials by varying the specific interaction rules, it includes natural methods for adjusting the computational resource demands by varying the number of particles and complexity of their interactions, and it is relatively simple compared to alternative physical simulation methods. However, while real liquids are much more complex than rigid solids, a main contribution of the present paper is to demonstrate that highly approximate methods for simulating fluids may still be sufficient for the every-day kinds of predictions that people make, and that these approximations may still be cognitively and computationally plausible. Particles are an intuitive idea for simulating physical systems that is useful across a range of physical scales. Real fluids are composed of on the order of $10^{23}$ particles (molecules) per kilogram. Physics, engineering, and graphics applications may use on the order of $10^5$ up to $10^7$ particles to achieve sufficiently realistic simulations \cite{bender2011sph,akinci2013versatile}, depending on the specific application. Particle systems are also commonly employed in game engines that must support liquids and soft bodies alongside rigid bodies highly efficiently, to allow for user interactions in real time, even on small mobile devices. A small number of particles ($10^3$-$10^4$) can achieve sufficiently accurate and stable simulations for the purposes of many computer and mobile games. Here, by contrast, we show that human data is best-accounted with on the order of a hundred particles, $10^2$, or even fewer.  Thus brains may have evolved representations for intuitive simulation of fluids similar to those used in contemporary computer simulations, but trading off accuracy for computational efficiency to an extreme that goes well beyond even the most severe short-cuts taken in conventional engineering applications. 
 
\paragraph*{Background of related research.}  We study the judgments people make about complex physical dynamics -- liquid flow -- in rich, realistically rendered scenes. This is very different from the highly simplified and abstract stimuli studied in some of the classic work on intuitive psychology in the cognitive psychology literature. Also, our focus on people's ability to make sophisticated quantitative physical predictions in complex everyday scenes \cite{battaglia2013simulation,hamrick2016inferring} may appear incongruous next to classic studies of intuitive physics that emphasized ways in which people's judgments about simple scenarios of objects in motion can be inconsistent with even very basic principles of Newtonian mechanics \cite{mccloskey1983naive}. In part this is just a difference in emphasis: We are interested in explaining how people make  successful predictions in complex everyday interactive settings, capacities that no robot, computer vision or AI system currently comes close to, rather than the mistakes people make in much simpler settings which are important for formal physics education but not necessarily for everyday physical interaction with the world. It is important to note, however, that even in previous work which emphasized the failures of people's physical intuitions, it was often the case that a majority or plurality of participants' judgments in many of the classic studies were consistent with Newtonian principles or approximate Newtonian simulations \cite{mccloskey1983intuitive,mccloskey1983naive,kaiser1986intuitive,cook1994constructing}; it is just the striking patterns of error which occurred in some cases that were presented as the main findings. With some exceptions \cite{mccloskey1983intuitive}, the classic literature often focused on textbook-style problems, typically presented with static, illustrated diagrams, and requiring participants to draw out future trajectories on paper. In contrast, recent literature has focused on more perceptual and perceptual-motor judgments, typically with rich dynamic displays and implicit inferences about what will happen next over short time scales (but see also, e.g. \cite{kaiser1992influence} for older examples). Recently, \cite{smith2013consistent} even replicated \cite{caramazza1981naive}'s results showing that people's drawings of predicted ballistic motion are often very inaccurate, but found that when those same participants were tasked with an interactive task of intercepting a moving object or releasing it to fall with a desired trajectory, their predictions were consistent with Newtonian dynamics. Most relevant to the present work, \cite{kaiser1986intuitive} found that subjects were better at judging the trajectory of water exiting a hose than a ball exiting a curved tube, hinting toward a crucial difference in physical predictions with fluids when people can access representations they have developed from their everyday interactions. 

These and other findings \cite{zago2005cognitive} are beginning to coalesce on an explanation of why people's physical intuitions appear to vary greatly in their accuracy: higher-level, deliberate reasoning about physics differs markedly from physical knowledge available to lower-level perceptual and motor systems. We speculate that higher-level systems, which are more flexible, not specific to physical reasoning, and based more on learning, may not benefit from the perceptual conditions and reward incentives that give rise to accurate physical knowledge in lower-level systems. Hence, our underlying hypothesis is that there are multiple systems supporting physical scene understanding, one of which is physical simulation. The purpose of this work is not to explore the boundaries between the different cognitive mechanisms for understanding physical systems, but rather to provide the first examination of mental simulation as an account of people's ability to predict fluid flow in realistic situations, and to develop the first computational models of these simulation abilities that can capture to a reasonable first approximation how people might make these predictions. 

Previous artificial intelligence research on reasoning about fluids has often pursued qualitative \cite{forbus2011qualitative,davis2008liquids,kim1990qualitative,kim1993qualitative,collins1987reasoning} or logical \cite{hayes1978naive, davis2008liquids} approaches, which are computationally efficient, and were developed prior to more recent hardware and algorithmic advances that support fluid simulation in real-time on mobile devices. Qualitative and logical models have been especially useful in situations that ask for explicit reasoning about a range of physical systems that people do not have much direct perceptual-motor experience with (such as thermodynamic or hydrodynamic systems or electrical networks).  More recently, \cite{davis2008liquids} developed a system of reasoning using first-order logic to draw conclusions about transferring liquid from one container to another in theoretical, two-dimensional scenarios. A key advantage of this approach is that it can be applied even when no numerical information about container geometries, relative distances, initial conditions or volume of the water, pouring speed, etc. is available. At the same time, however, it is unclear how these qualitative or logical models could account for the fine-grained quantitative judgments that participants make in the perceptual settings we study here.  A final class of classic AI approaches to intuitive physics is analogical simulation, e.g., \cite{gardin1989analogical} devised an analogical particle-based fluid simulator, whose particles ``communicate'' with each other based on a set of eight rules (see Fig~\ref{fig:heuristic}C). This approach combines some aspects of qualitative reasoning and quantitative, physics-engine-based simulations, and can be seen as a simpler, more heuristic version of physically based simulations. Most recently, deep neural networks have been proposed as a very different approach to making physical predictions on short time scales \cite{lerer2016learning}.  Both analogical simulation and neural network approaches can be applied to our tasks, and we evaluate simple versions of both on our experimental data, in addition to the more quantitative approximate simulation models we have focused on. 

Proponents of qualitative approaches have also argued that solving problems that involve predicting fluid motion ``to a high degree of accuracy involves computational fluid dynamics'' and ``it is quite unlikely that we are capable of performing such a prodigious feat mentally'' \cite{forbus1997qualitative}. But it is still unclear to what extent qualitative approaches can acount for people's sophisticated capacity for understanding and interacting with a wide variety of fluids and fluid-like materials in everyday settings. Furthermore, there are strong reasons to believe fluid simulation in the brain is not prohibitively expensive. Particle simulations can be run in parallel on GPUs, and as we argue here, the brain may be able to get by with just a coarse approximation for its purposes. Such simulations require computation that is on par (or less) than that used by today's deep neural networks for computer vision \cite{lecun2015deep}, which are thought to be reasonable models of parts of biological visual processing \cite{yamins2014performance}. In addition, \cite{battaglia2016interaction}'s recent "interaction networks" and \cite{chang2016compositional}'s "neural physics engine" models introduce neural networks which operate on objects and relations, and can learn very accurate quantitative physical simulators from data. In light of this, as well as the mounting evidence that human predictions and inferences about rigid physical objects is supported by simulation, we believe it is plausible that humans perform quantitative simulations, and not justifiable to dismiss this possibility out of hand. 

Some psychological work in vision has looked at how humans perceive liquid viscosity, and has also been skeptical of the sort of computations we propose. For example, \cite{kawabe2015seeing, paulun2015seeing} argue that perception of liquids relies on superficial cues, and show that people use certain statistical motion cues to predict viscosity. Furthermore, they argue that, ``Given that the liquid image motion is a result of...complex physical processes, it seems practically impossible for the visual system to infer the underlying physical movements of particles from observed movements using inverse optics computations." We do not disagree that surface appearance cues can be important in how people perceive liquid properties.  However, such cues alone cannot directly explain how people can \textit{predict} the motion of liquids, which is the focus of our experimental work.  

The remainder of this paper is organized as follows. We first present our simulation model and the particle-based simulation method it uses, followed by several alternative models, including simpler simulation alternatives, as well as deep neural networks. We then present Experiment 1, which demonstrates that participants are coarsely able to predict the complex physics of both scenes involving low-viscosity liquids (e.g., water) and scenes involving high-viscosity liquids (e.g., honey), and explores the roles of damping (see Intuitive fluids engine) and uncertainty in our model's correspondence to participants' judgments. Experiment 2 then focuses more directly on people's sensitivity to viscosity. Finally, we discuss implications of both experiments and directions for future work.

\section*{Materials and methods}
\subsection*{Models}
\subsubsection*{Physical simulation}

\paragraph*{Smoothed-particle hydrodynamics.} Our computational cognitive model is built on a particle-based approach to simulating fluids. We define \emph{simulation} as a process that applies fixed rules iteratively, over a sequence of steps, to approximately predict a system's state over time. The specifics of the simulation algorithm in our model is not crucial to our theory; it implements the general principles that a fluid's density is approximated as particles, that dynamics of the fluid correspond to the approximately Newtonian dynamics of the individual particles, that the fluid properties (e.g., viscosity, stickiness, etc.) are distinguished by different rules for how the particles interact, and that prediction precision and computational resource demands trade off by varying the numbers of particles, temporal resolution of the simulated time steps, and complexity of the particle dynamics and interaction rules. 

\begin{figure}[h]
\centering
\includegraphics[width=0.75\linewidth]{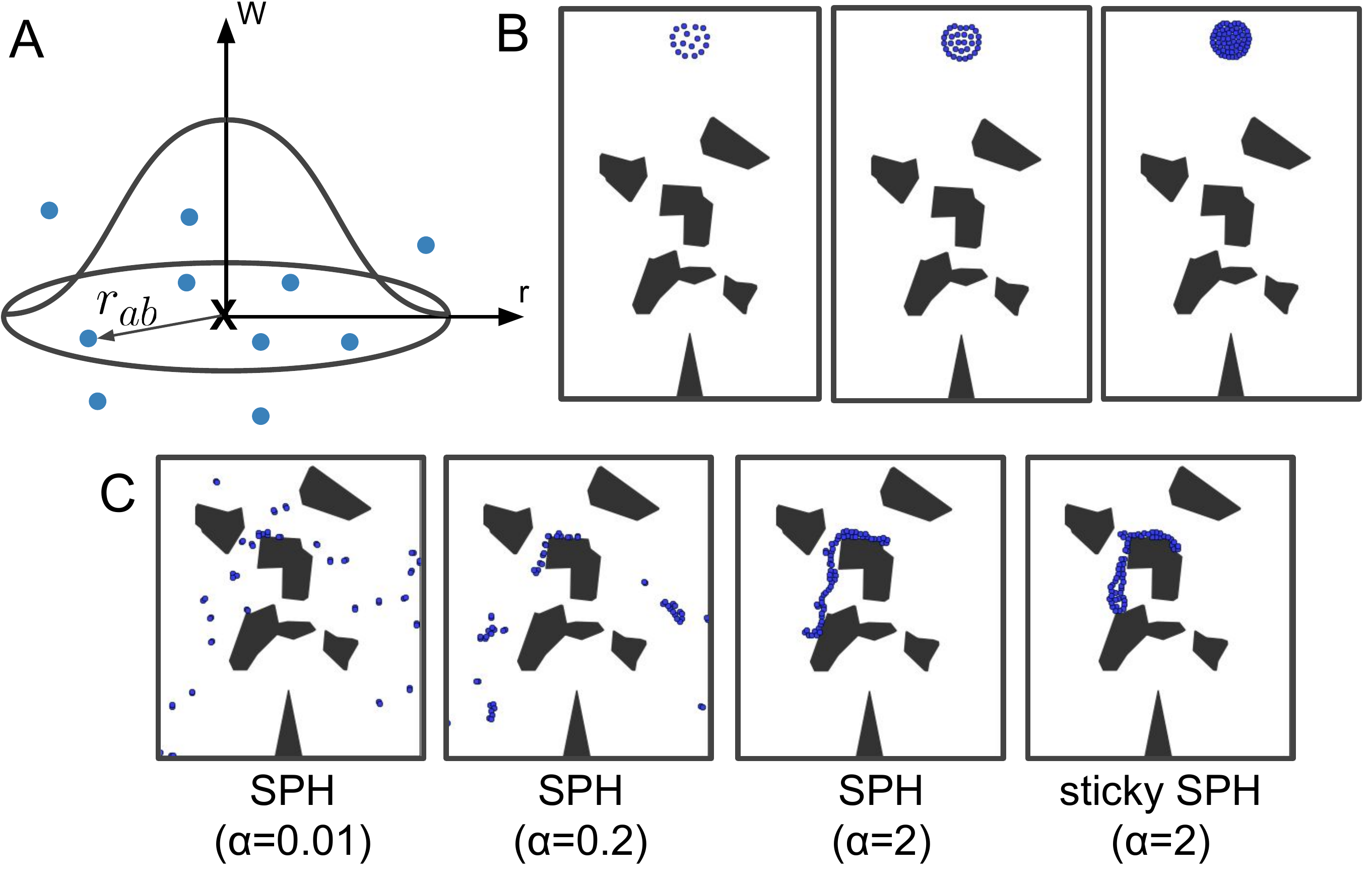}
\caption{\label{fig:model} \small SPH overview. (\textit{A}) How SPH approximates a fluid. For any location in the fluid, marked ``X'' on the diagram, particles in the local neighborhood are used to approximate the fluid's, density, pressure, and dynamics at that point. The bell-shaped envelope depicts the strength of each neighbor's influence on the approximation, which falls off with distance. 
(\textit{B}) SPH simulations can be allocated more resources to achieve more precise approximations. In the second and third panels, more particles are allocated than in the first, which will result in more accurate and stable simulated fluid dynamics. (\textit{C}) The rules by which particles interact can be varied to produce different qualitative fluids and materials. The first three panels show differences in splashing behavior as a function of viscosity. The fourth panel shows a non-Newtonian fluid that sticks to rigid surfaces (like honey).}
\end{figure}

There are various classes of fluid simulators one could choose, such as volumetric, particle-based, or more qualitative, which all have in common the property of a time step. Within each broad class, there are various possible implementations. As a starting point, here we choose to explore one particular particle-based method. One possible concern regarding the cognitive plausibility of fluid simulation models is the choice of time-step resolution. The particular implementation we use requires a relatively fine time step for stability (0.0002 seconds). But others have implemented stable simulations with the same method using a substantially larger time step of 0.01 seconds, which is sufficient for real-time user interaction with hundreds of particles on a mobile device. Other particle-based simulation methods achieve stability with similarly large time steps \cite{macklin2014unified}. 

Here we use smoothed-particle hydrodynamics (SPH) \cite{monaghan2005smoothed}, a method from computational fluid dynamics, as the core algorithm for our model. SPH is used widely in graphics and video games for approximating the dynamics of many types of compressible and incompressible fluids (e.g., liquids and gases). The state of the fluid is represented by a set of particles at discrete time steps. Each particle carries information about a volume of fluid in a particular locality in space, including its position, velocity, density, pressure, and mass. On each simulation time step, the particles' densities and pressures are computed, which are then used to update the accelerations, velocities and positions. A particle's density is calculated by interpolating its neighbors' densities, weighted by their distances, 
$\rho_i=\sum\limits_{j=1}^{N_i} m W(r_{ij},h)$, where $\rho_i$ is the density at particle $i$'s location, $m$ is the mass of each particle, $W$ is the kernel function, and $r_{ij}$ is the distance between particles $i$ and $j$ (Fig~\ref{fig:model}A). The weighting is determined by $W$, which has a cutoff radius, $h$, beyond which particles have no influence. After computing particle $i$'s density, its pressure is updated, followed by particle-particle friction damping forces (analogous to viscosity). Its acceleration is a linear combination of the pressure and damping, the velocity update is proportional to the acceleration, and the position update is proportional to the velocity. 

The precision of the liquid simulation can be adjusted by how many particles are used: with more particles, the simulated liquid's movement is more closely matched to that of a real liquid (Fig~\ref{fig:model}B). But increasing the number of particles also increases the computational cost of the simulation, thus effecting a trade-off between efficiency and accuracy.\\ 

\paragraph*{Intuitive fluids engine.} Our simulation-based cognitive model, which we term the ``intuitive fluids engine" (IFE), is analogous to \cite{battaglia2013simulation}'s intuitive physics engine (IPE), but is capable of predicting a fluid's dynamics. It posits that when the brain observes the initial conditions of a physical scene that contains fluids, it instantiates a corresponding particle-based simulation (i.e., SPH) to predict future states of the scene. SPH comprises a family of related particle-based algorithms, and our implementation is similar to the most popular versions. Our model takes as input the configuration of the scene, including the solid elements and the fluid's spatial state and material attributes, such as particle friction ($\alpha$) and particle stickiness. The number of particles, $N$, that are instantiated can be varied as a means of adjusting the computational resources allocated to the simulation. 

Inconsistency between people's mental simulations and ground truth physics (or our model's approximation thereof) may come from two different sources: perceptual uncertainty and incorrect physical assumptions. We explored both of these possibilities in our model. Our task required predicting how water will splash, falling under the full acceleration of gravity. Such splashes are energetic and happen on small time scales, and therefore it should be difficult for people to predict their details exactly. We suspect that people may somehow ``slow down'' the fluid in order to try to predict splashes in more detail, or may simply underestimate the potential energy of the liquid. We explored this possibility by incorporating a damping term into the IFE simulations. This term decreases the acceleration in each particle at every time step by a fixed proportion, $\zeta$, of that particle's current velocity, resulting in a liquid that falls more slowly and splashes less energetically. Values of $\zeta$ were discretized to a spacing of $1.0$ (i.e. we simulated 21 different values of $\zeta$ between $\zeta=0$ and $\zeta=20$). 

In prior work \cite{bateshumans}, we explored a model of perceptual uncertainty that randomly perturbed the initial position of the fluid and averaged the outcome over many random draws, but did not consider potentially incorrect physical assumptions. Here our model's uncertainty is implemented by averaging the predictions made over a small range of damping values (where $\sigma$ represents the size of the range). Thus, $\sigma=0$ represents a deterministic IFE and $\sigma \neq 0$ represents a probabilistic IFE.

In our simulations of the stimuli, a liquid was initially positioned at the top of the scene, then moved downward under the force of gravity through a set of solid obstacles, and accumulated in one or more containers at the bottom. In Experiment 1, the bottom was divided into two equally sized basins, left and right. The model's judgments, $J$, took values between $0$ and $1$, where $0$ represented all liquid flowing into the left basin, and 1 represented all flowing into the right, i.e.: $J = n_{right} / N$, where $N$ was the total number of particles, and $n_{right}$ was the number that flowed into the right basin. In Experiment 2, there was a cup positioned at the bottom of the scene and $J=1$ represented all the liquid accumulating inside the cup, while $J<1$ indicated some liquid collected in the bottom of the scene, outside of the cup. We also computed predictions from a ``ground truth'' model which did not include uncertainty or damping, and which used a single, deterministic simulation with a high number of particles and the correct viscosity, to predict the liquid behavior as accurately as possible. We emphasize, however, that what we call ground truth here is still an approximation to true physics, but should in theory be more accurate.

We created different SPH liquids with particle friction values that ranged from low (approximating water) to high (approximating honey). However, high viscosity Newtonian liquids behave differently than real honey, because they do not stick to surfaces (SPH particles collide with obstacle surfaces as inelastic spheres, while real fluids have much more complex boundary interactions). Fig~\ref{fig:model}C reports particle friction values, $\alpha$, for various liquids. Our $\alpha$ corresponds to that in the artificial viscosity term of \cite{monaghan1992smoothed}. In order to model non-Newtonian, sticky liquids, we created a variant of the basic SPH liquid described so far, which implements particle stickiness by damping the normal and parallel components of velocity for particles that are in collision with solid obstacles (Fig~\ref{fig:model}C, rightmost panel), where `normal' and `parallel' are with respect to the obstacles surface. Due to the computational demands of running a wide range of fluid simulations, we only considered two sets of sticky parameter values, chosen by hand \textit{a priori} to qualitatively match the visual behavior of the stimuli as closely as possible (details below). The values for the two stickiness parameters varied slightly between Experiment 1 and 2, because in Experiment 2, we made the honey somewhat less viscous than in Experiment 1. The space of possible sticky SPH liquids could in theory be further explored by varying the viscosity in combination with the two parameters controlling the velocity damping. In fact, the full parameter space could be considered as having five parameters ($N$, $\zeta$, $\alpha$, and normal and parallel components of damping during collision), but note that adding stickiness to low-viscosity fluids makes little physical sense, as the particles can slide easily past each other compared to more viscous fluids. The result would be largely the same as water, but there would be a layer of particles left stuck to obstacles. This behavior bears little resemblance to any known liquid. Thus, we deem the chosen fixed values to be a sufficient starting point, and distinguish sticky and non-sticky liquids as separate models for the purposes of presentation. All analysis below involving the sticky liquid ("Model 1 honey"; see below) will use these fixed parameter values for $\alpha$ and stickiness.

The above model parameters fall into two categories: those that control the fidelity of simulation ($\sigma$, $N$) and those that determine the physics of the liquid ($\zeta$, $\alpha$, particle stickiness). Settings for these parameters are discussed below, in Intuitive fluids engine parameters.

\paragraph*{MarbleSim.} We explored another physically-based, but computationally less-demanding simulator, which was akin to representing the fluid as a handful of marbles. This simulator, which we call ``MarbleSim'', is identical to the IFE in that it instantiates particles to represent the fluid. However, MarbleSim replaces the SPH forces with rigid-body interaction rules that are standard in rigid-body physics engines. This reduces computational complexity (and simulation time), because there are many fewer pairwise interactions to calculate between particles. In MarbleSim, each particle is represented as a solid sphere that collides inelastically with obstacles and other particles. The IFE simulator also instantiates particles as rigid spheres, but is different in that it ignores particle-particle collisions, allowing them to pass through each other. (Note, however, that the SPH forces will generally prevent particles from getting arbitrarily close.) Thus, particle-obstacle interactions are identical in both the IFE and MarbleSim, but particle-particle interactions in MarbleSim do not involve SPH forces--particles instead collide with each other as rigid bodies, similarly to the way marbles would. All collisions are perfectly inelastic (i.e. particles do not bounce) in both MarbleSim and the IFE, as this is most consistent with the behavior of liquids.

While particles in MarbleSim have a friction parameter that controls how easily they slide past each other when they touch, this is not analogous to the particle friction in the IFE ($\alpha$). Furthermore, this parameter has very little effect on the simulation outcomes in our setting. In addition, stickiness, which we employ in SPH simulations, would make little physical sense in MarbleSim, since the particles easily roll past each other, rather than cohering together as a continuous substance. The result is qualitatively similar in behavior to simulations without stickiness, but varies in outcome because of the layer of stuck particles ``left behind'' on the obstacles. Thus, similar to the case of sticky, low-viscosity liquids addressed above, a ``sticky marble'' simulation bears little resemblance to the behavior of either water or honey. However, MarbleSim was still allowed to vary along two dimensions, $N$ and $\zeta$, which exactly match the corresponding parameters in the IFE, and were allowed to vary along the same ranges of values. Also, we found that performance of the model improved when adding uncertainty over damping, as we did with the IFE. The results reported below use the same value for $\sigma$ as the uncertain IFE.

\subsubsection*{Heuristic simulation}
\begin{figure}[h]
\centering
\includegraphics[width=1.0\linewidth]{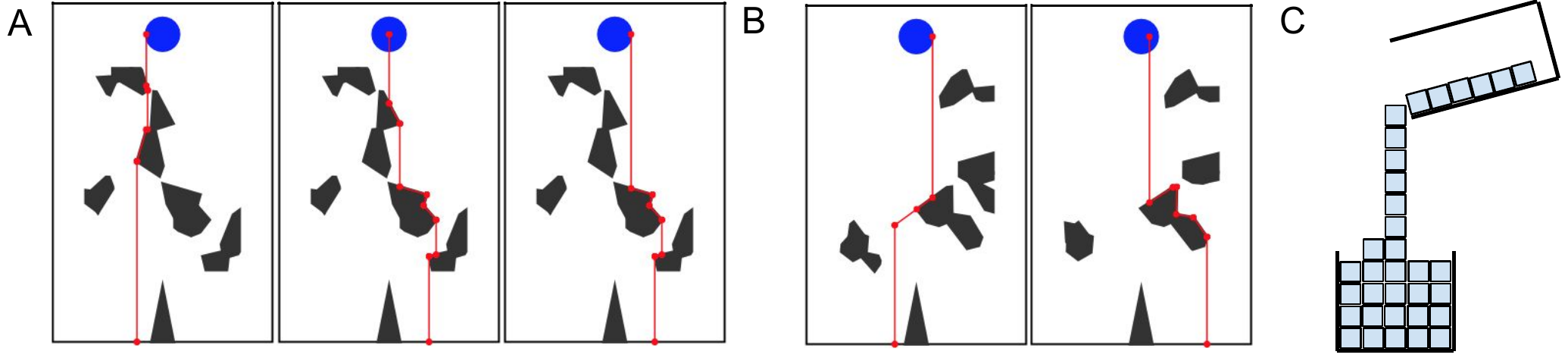}
\caption{\label{fig:heuristic} \small Gravity heuristic. (\textit{A}) Each panel depicts the path of a different particle. (\textit{B}) Depiction of `momentum' (left) for $m=0.1$, and `global noise' with $g\neq 0$ (right), for a single particle that has chosen the ``wrong'' direction (going against gravity). See text for details. (\textit{C}) Visualization of model from Gardin and Meltzer (1989). The simulated fluid (blue cubes) travels straight down under gravity, which is analogous to \textit{A}.
}
\end{figure}

\paragraph*{SimpleSim.} We also explored a simpler kind of dynamic simulation, which is comparable to the analogical models of \cite{gardin1989analogical} in the way it uses a set of deterministic, recursive heuristics to form predictions.  These heuristic rules are inspired by physics but are not in any formal correspondence to physical dynamics.  We add to these rules notions of noise and momentum to produce a range of different ``fluids''.  This model class, which we call ``SimpleSim'', instantiates equally-spaced ``particles'' along the midline of the liquid's starting position and generates a path straight downward (in the direction of gravity). In the model's simplest form, when an obstacle is encountered, the path continues along the obstacle surface until it can go straight down again. The direction to travel along the obstacle's surface (left or right) is chosen to follow the local gradient downward, with gravity (see Fig~\ref{fig:heuristic}A). For example, if the polygon edge that the particle collides with slopes down and to the right, the particle trajectory continues down and to the right along the surface. However, the model also includes two degrees of freedom that allow it to deviate from this basic behavior. The first parameter is a `global' noise parameter, $g$ (ranging from 0 to 1), that controls the probability of choosing the ``wrong'' direction when a particle encounters an obstacle, such that the particle travels initially travels against gravity. For example, if the colliding edge has a negative slope (down and to the right), the model may randomly choose to continue up and to the left instead of down and to the right (see Fig~\ref{fig:heuristic}B, right panel). A particle makes this decision about whether to initially travel with or against gravity once per obstacle, upon initial collision, and then continues in that same direction until it reaches the obstacle's edge. The second parameter, $m$ (ranging from 0 to 0.2), is somewhat akin to momentum: a particle does not immediately drop straight down upon reaching the edge of an obstacle. Rather, it continues for a fixed distance in the same direction it was traveling, before dropping straight down (see Fig~\ref{fig:heuristic}B, left panel). The parameter value specifies the magnitude of this distance. Intuitively, when $m$ is smaller, the model should make predictions closer to honey, which clings more to the obstacles, and when it is larger, the model should make predictions closer to water, which sloshes off with some momentum. Both parameters were discretized to 11 evenly-spaced values, and thus we calculated predictions for 121 total parameter settings. The value of $m$ corresponded to the extra distance (in meters) traveled in the horizontal direction (orthogonal to gravity). We chose $m$ to correspond to the horizontal distance, instead of absolute distance, in order to mirror real Newtonian physics, which separates vector quantities, such as momentum, into orthogonal components. Finally, we found that adding additional uncertainty over $m$, in the same way as we did for the IFE and MarbleSim, modestly improved fits with the data. See Results for details.

The model's predictions were calculated by counting the number of ``particles'' that end up on each side of the divider (or in the cup, for Experiment 2), as in the simulation models. The model reported here used 100 particles, but the results do not depend crucially on the exact number of particles, as the particles' trajectories are completely independent and do not interfere with each other.

\subsubsection*{Models without dynamics}
We contrasted the simulation-based models above with two alternatives that did not explicitly seek to simulate dynamic fluid motion: one (the ``gravity heuristic'') that is still in a sense a simulation model, but which runs a purely geometric simulation, embodying the heuristic notion that gravity pulls all liquid particles down unless an object blocks their path; and another that used a deep convolutional neural network~\cite{krizhevsky2012imagenet,jia2014caffe} trained on thousands of examples similar to our experimental test conditions to predict the task target as a pattern recognition problem.  

Both of these alternatives are highly specialized to the particular task studied, and are unlikely to generalize to liquid motion in even slightly different conditions (e.g., different obstacle arrangements, different liquid properties), in contrast to the models above which ought to generalize whenever the underlying physics being modeled is shared. However, they represent reasonable and well-known competing perspectives on the general mechanisms of human perception and cognition \cite{gigerenzer2011heuristics, mcclelland2013integrating}, and offer advantages such as simplicity of computation and straightforward statements about the learning processes.

\paragraph*{Gravity heuristic.} Formally this model corresponds simply to the SimpleSim model with $m=0$ and $g=0$. With these parameters set to null values, the model simply follows the geometry of the obstacles, and it makes the same predictions regardless of the liquid type. Thus, it may be considered a simple, geometric heuristic, formalizing the prediction that particles should continue to ``go down'' under gravity, without penetrating obstacles or showing any momentum and the dynamics that come from that. 

\paragraph*{Convolutional networks.} As an alternative to explicit physical representations of liquid dynamics, we also tested the possibility that people might be responding based on a simpler mapping between visual input and physical outcome, learned from experience using purely visual features of static scenes. Our models are intended to capture only quick, bottom-up, visually guided intuitions that people might have developed from experience. We implemented a deep learning model, replacing the top layer of a widely used convolutional neural network~\cite{krizhevsky2012imagenet}, pre-trained on a very large collection of images, with a linear output layer, and performing backpropagation~\cite{jia2014caffe} until convergence to learn a regression from images to labels in a supervised fashion. We trained separate networks for water and honey outcomes in each experiment, and refer to each as ``water ConvNet'' and ``honey ConvNet'', respectively. The dataset for the water ConvNet in each experiment consisted of 10,000 randomly generated scenes, with target values in the range $[0,1]$ that corresponded to the proportion of water that went to the right bin in Experiment 1, or in the cup in Experiment 2 (as determined by a deterministic ground truth simulation with $N=100$). The network was not shown any intermediate fluid positions. The honey ConvNet in each experiment was trained by fine-tuning the water model using $\sim$2500 training examples. All networks were trained in layer-wise stages to minimize overfitting, freezing all layers below layer \textit{n}, for $n= [N, \ldots, 1]$, where $N$ is the total number of layers. That is, the first training stage freezes all but the last layer, the second stage freezes all but the last two layers, etc. In addition, we augmented our training data with versions of the original images zoomed by random amounts.

\subsection*{Experiment 1}

\subsubsection*{Participants}
All participants (N=65) were recruited from MIT Brain and Cognitive Sciences' human participants database (composed of roughly half MIT students and employees, and half local community members). All gave informed consent, were treated according to protocol approved by MIT's IRB, and were compensated \$10/h for participation. All experimental sessions were one hour long, and each participant ran in one session in one experiment. All had normal or corrected-to-normal vision. Stimuli were presented on a liquid-crystal display, which participants free-viewed from a distance of 0.5-0.75 m. They indicated their responses by depressing a key on the keyboard, or by adjusting the computer mouse and then clicking to lock in their choice. 

\subsubsection*{Stimuli and procedure}
In order to test people's ability to predict the behavior of a liquid, participants were presented with 120 virtual scenes, simulated in 3 dimensions (including depth into the screen), with a simulated size of 1.0 m x 1.50 m. The scenes depicted a cylindrical volume (diameter equal to 0.148 m) of liquid positioned above a randomly generated obstacle course composed of fixed, solid objects and asked to predict what fraction of the liquid would flow into each basin below the obstacles under gravity. Participants were randomly assigned to one of two groups: one group was presented with a low-viscosity, water-like liquid (Experiment 1a), the other with a high-viscosity, honey-like liquid (Experiment 1b). The scenes' geometries were automatically generated as described below. During practice trials (see below), participants viewed videos of the simulated liquid falling under gravity, rendered at 30 Hz using Blender's (www.blender.org) Cycles ray-tracer. All moving liquid videos were simulated using Blender's Lattice-Boltzmann liquid simulator (see Appendix A for simulation details). Lattice-Boltzmann methods are very different from SPH, a choice intended to eliminate the possibility that our model would only capture people's judgments because the scenes were generated using a similar fluid simulation algorithm. All participants were presented with the same, randomly shuffled scene order. The 120 trials were divided into three blocks of 40, with a short break in between. Both Experiments 1a and 1b included a practice and test phase. 

On each trial (in both practice and test phases), participants viewed an image of the scene, with the liquid in its starting position. They were instructed to predict the proportion of liquid that would end up on each side of the divider (see the dark wedge at the bottom of the stimuli in Fig~\ref{fig:stimuli}A), and to indicate their judgment by moving a virtual slider with the mouse left or right, then pressing ENTER to submit. [See online supplementary material for stimuli videos.]

During the practice phase, comprised of 15 trials, participants received visual feedback after submitting their response on each trial. During feedback, subjects saw a rendered video of the liquid flowing through the obstacles (4 seconds for water and 10 seconds for honey; made in Blender, as described above). The practice phase was designed to familiarize participants with the characteristics of the liquid and the response procedure.

\paragraph*{Random scene generation}
The obstacles in the test scenes were generated automatically by first dividing a plane into polygonal cells using 2D Voronoi tessellation, then selecting a random subset as solid obstacles such that the total area approximately summed to 0.12 m\textsuperscript{2}. Coarse SPH simulations were run to filter out those scenes in which liquid particles remained trapped in obstacle concavities or had little interaction with the obstacles.

\paragraph*{Intuitive fluids engine parameters}
We partition the parameter space into three distinct classes of liquids--non-sticky (``water''), sticky (``Model 1 honey''), and highly-damped (``Model 2 honey''; see below). The water and honey classes correspond to the liquids in Experiments 1a and 1b, respectively. For the water IFE we explored particle friction values of $\alpha=\{0.01, 0.2, 2\}$. The Model 1 honey IFE used $\alpha=1.25$ for Experiment 1 and $\alpha=2$ for Experiment 2. (As mentioned above, the Model 1 honey liquids also involved two extra parameters controlling stickiness, which were fixed \textit{a priori} to qualitatively match the stimuli in honey trials of each experiment, respectively.) In total, the IFE has six sub-models: water, Model 1 honey, Model 2 honey, which could all be either deterministic or uncertain (by making $\sigma$ zero or non-zero). However, we only report results from the uncertain versions of these three fluid types, as they had slightly higher correlations with the data across most of parameter space, but maintained the same qualitative trends.

The number of particles used to represent the liquid was varied from 1 to 100. However, with fewer than 15 particles in our settings, the assumptions of SPH are violated to an extent that it cannot be considered to be a fluid simulation. We explored simulations with fewer particles in order to compare their predictions with people, and found them to have poor fits to the data, and therefore we will not report those results. With greater than 100 particles, non-sticky simulations converge to very similar outcomes. The model's damping parameter was varied from $\zeta=0$ (no damping) to $\zeta=20$ (high damping). The value of $\sigma$ was $0$ (i.e. no uncertainty) for the deterministic IFE, and $4$ for the uncertain IFE (but results were similar for other values). To compute predictions in the uncertain IFE for a particular value of damping, $\zeta$, deterministic simulation outcomes were averaged over damping values in the range $[\zeta-\sigma, \zeta+\sigma]$. Each model (deterministic and uncertain IFEs) made a single prediction for each scene and unique setting of fluid parameter values ($N$, $\alpha$ and particle stickiness, $\zeta$). The ground truth models corresponded to $\zeta=0$, $\sigma=0$, and $N=100$ (for honey) or $N=200$ (for water). The number of particles for honey ground truth was fewer than for water because as the number of particles is increased, the fluid behaves as if it were non-sticky. This is because only the particles in contact with an obstacle have modified behavior, and so their impact on the overall dynamics of the fluid diminishes as they become a smaller proportion of the total number of particles. We chose $N=100$ for ground truth, because during our search for a liquid that qualitatively matched honey from each of the experiments, this is the number of particles we used. Thus, it is the closest match we could find to the behavior of the stimulus fluid. Further work should be dedicated to developing improved simulation methods to more accurately capture the dynamics of sticky liquids, in a way that is independent of number of particles.

As mentioned above, we explored two alternative models to account for judgments about honey stimuli. The first (Model 1) simulated honey as a high particle-friction and particle-stickiness fluid, as described above. The second alternative (Model 2) simply corresponded to the water IFE with high damping. Thus, Model 2 can be considered a subspace of the water IFE. It was found empirically that these two models predict similar outcomes in our tasks, and so we consider each as an alternative cognitive hypothesis. Model 2 honey may be especially appealing from the standpoint of representational simplicity, as it replaces multiple parameters related to stickiness and viscosity with a single parameter, damping.

To avoid over-fitting, in general, we chose parameters based on Experiment 1, and fixed them to these values for Experiment 2 (see Table~\ref{tab:parameters_summary}). Further, we fixed our parameters in a stratified manner, starting by finding the best value for $N$, then fixing $N$ to this value while searching for the best values for $\alpha$ and $\zeta$. More specifically, to choose $N$, we took the average between the best-fitting number of particles for Experiment 1 water trials (with the water IFE) and Experiment 1 honey trials (with the Model 1 honey IFE), which was $N=50$. (Best-fitting values for water and honey trials independently were $N=25$ and $N=75$, respectively, but fits were similar across the range from $N=25$ to $N=75$ in both models.) With $N$ fixed, we then found the best-fit $\alpha$ value for the water IFE on Experiment 1a responses. Next, with $N$ and $\alpha$ fixed, we found the best $\zeta$ for the water IFE on Experiment 1a responses. All IFE models, across both experiments, shared the same $N$, as found above, and the water IFE in Experiment 2 inherited its parameters from Experiment 1. Model 1 honey inherited $N$ and $\zeta$ from the water IFE, but used the $\alpha$ that was appropriate for honey. Model 2 honey inherited $N$ and $\alpha$ from the water IFE, but fit $\zeta$ using responses from Experiment 1b. The final parameter choices for the water IFE were $N=50$, $\alpha=0.01$, and $\zeta=5$. The damping value for Model 2 honey was $\zeta=11$. Note that $\alpha=0.01$ was also the most physically accurate value of $\alpha$ for water. Finally, also note that there was no $\alpha$ value to fit for Model 1 honey, as it was not allowed to vary. 

\subsection*{Experiment 2}

\begin{figure}[tp]
\centering
\includegraphics[width=0.5\linewidth]{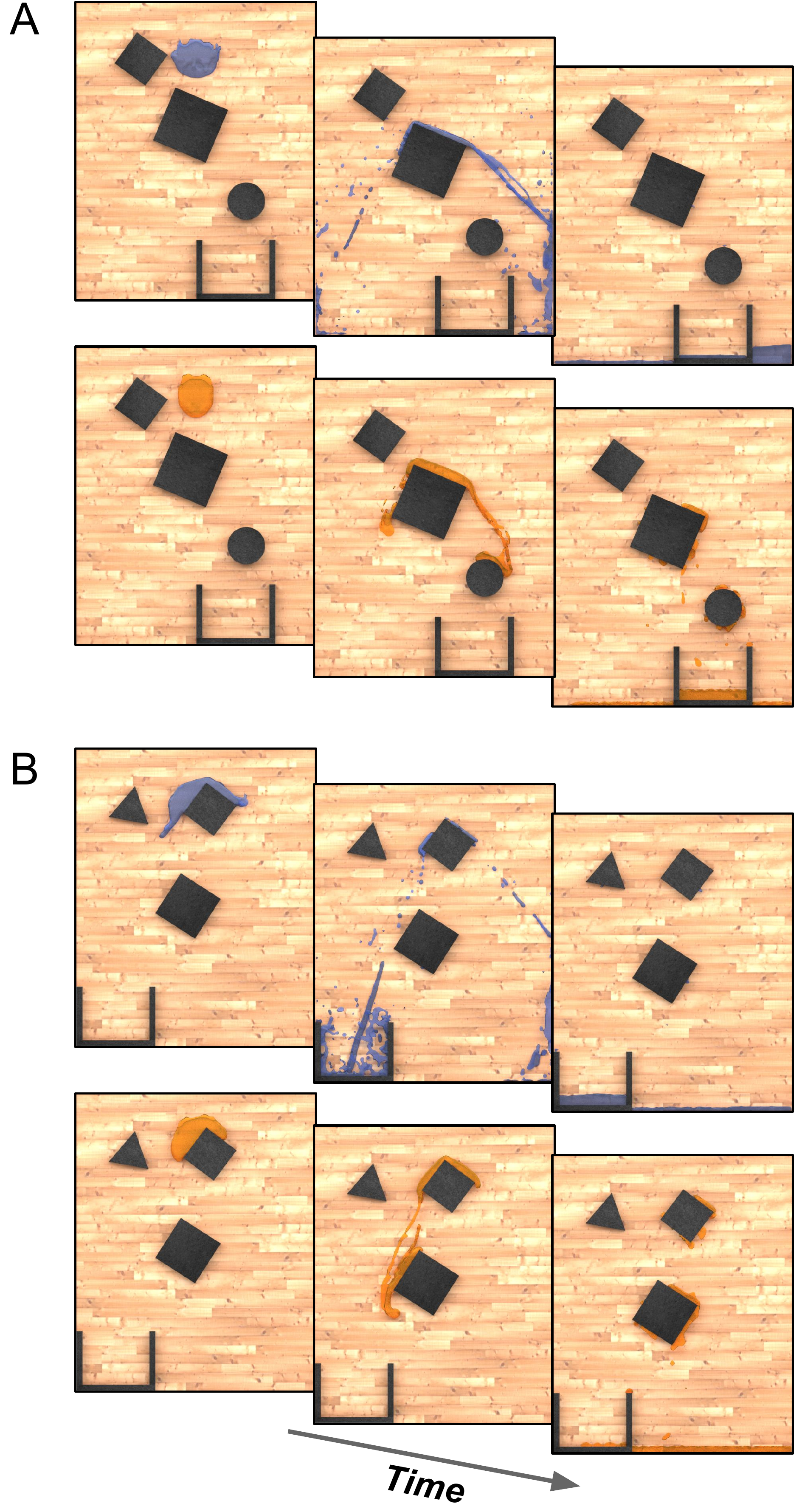}
\caption{\label{fig:motivate_exp2} \small Example scenes from Experiment 2 for which water and honey trials had very different outcomes. (\textit{A}) A scene in which much more honey flowed into the cup than water. (\textit{B}) A scene in which much more water flowed into the cup than honey.}
\end{figure}

For some model parameter settings in Experiment 1, both water and honey models made similar predictions, which limited the model's ability to assess people's sensitivity to viscosity and stickiness. In Experiment 2, we designed new stimuli to elicit highly different outcomes between low- and high-particle-friction and stickiness IFE simulations. See Fig~\ref{fig:motivate_exp2} for examples of Experiment 2's more strongly contrastive stimuli.

\subsubsection*{Stimuli and procedure}
We generated new random scenes and automatically selected scenes for which the model's predictions for honey versus water would be anti-correlated. This would allow us to more directly examine whether our model could explain people's sensitivities to the different liquids' physical properties. We also modified the task from Experiment 1, by removing the basin divider and instead placing a cup at the bottom of the scene (see Fig~\ref{fig:stimuli}B). Participants judged what percent of the liquid would end up in the cup, which was a third of the width of the scene, and could be at one of five evenly-spaced locations along the bottom. By varying the cup location and making the width less than the basins in Experiment 1, we could more easily create strongly anti-correlated scenes. Each scene consisted of 3 obstacles, which could either be an equilateral triangle, square, or circle, and varied randomly in size (within a specified range) and position. 

All 21 participants saw 120 trials (60 water and 60 honey). Each participant saw the same 60 scenes for each liquid type, but all honey trials were mirrored with respect to water, to ensure they weren't recognized. The honey stimulus in Experiment 2 had a lower viscosity than that of Experiment 1. (Experiment 1 honey was meant to be as different as possible from water, to try to elicit different participant responses. Experiment 2's honey was designed to be more typical of honey encountered in everyday situations, to draw on prior knowledge of how it behaves.) The honey model whose simulation parameters were most closely matched to the stimulus liquid also had a lower viscosity ($\alpha=1.25$) than that of Experiment 1 ($\alpha=2$).

\section*{Results}

\subsection*{Experiment 1}
We first asked: How accurate are participants' predictions about flowing liquids? In all analyses we calculated the means across participants' judgments for each scene, and estimated Pearson correlations between those mean judgments and each model's predictions as a measure of how well the model fit the human data. 

In Experiment 1a, the correlations between the mean participant responses and the water and honey ground truth models were $r=0.76[0.74, 0.78]$ and $r=0.83[0.81, 0.84]$, respectively (the interval in brackets is a $95\%$ CI, estimated by a bootstrap analysis with $10,000$ resamples \cite{efron1994introduction}). This is somewhat surprising because one would expect water simulations to better account for judgments about water than honey simulations (see Experiment 2). By contrast, Experiment 1b showed that participants' judgments about water and honey were best-explained by their corresponding ground truth models: $r=0.48[0.43, 0.53]$ for water ground truth and $r=0.73[0.69, 0.76]$ for honey ground truth. Fig~\ref{fig:double_dissociation} summarizes correlations for IFE parameter settings as described in the previous section.

\begin{figure}[h!]
\centering
\includegraphics[width=0.5\linewidth]{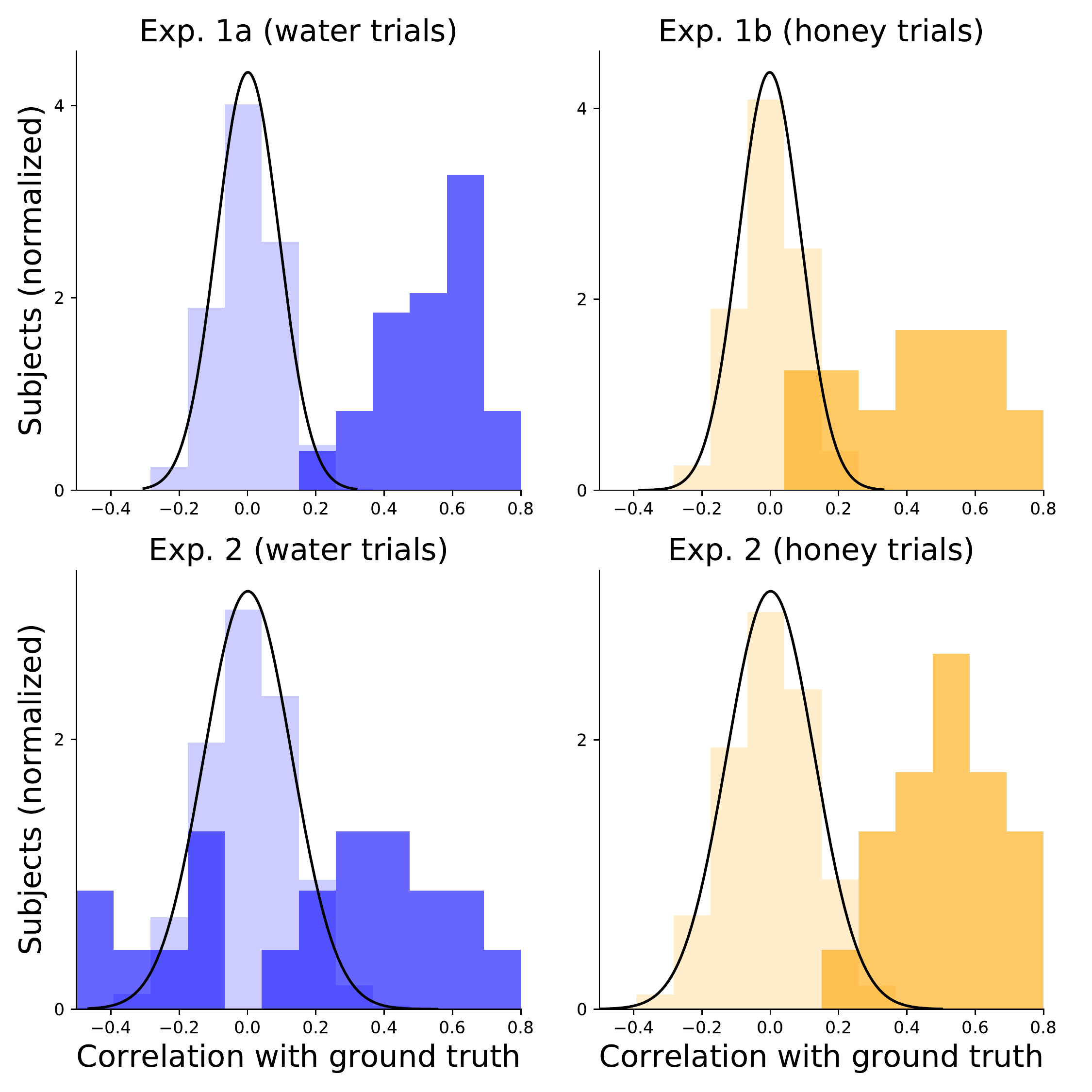}
\caption{\label{fig:indiv_vs_null_hist} \small Distribution of individual participant correlations with ground truth (dark colors) versus the null hypothesis (light colors).}
\end{figure}

\begin{figure}[tp]
\centering
\includegraphics[width=1\linewidth]{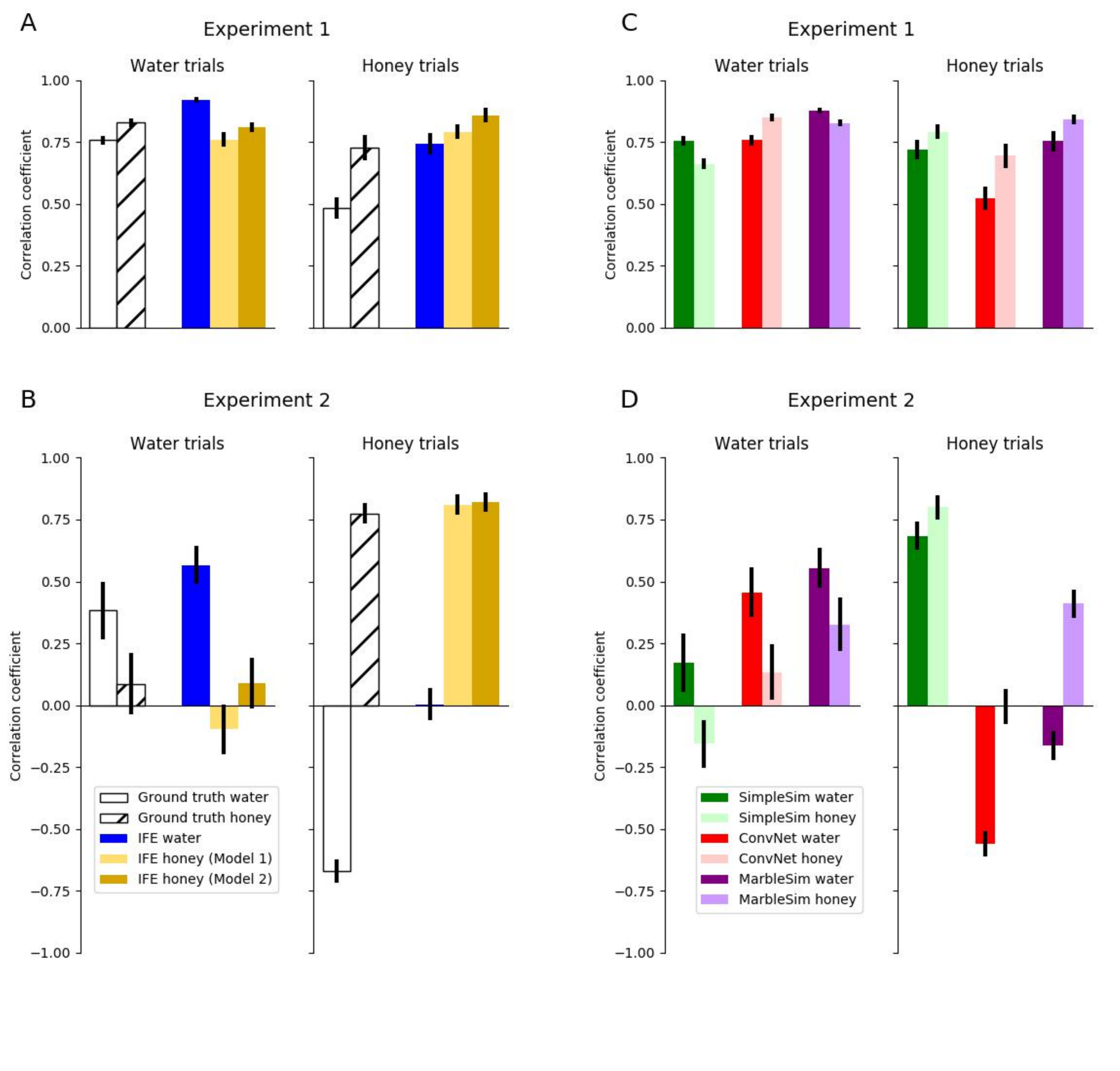}
\caption{\label{fig:double_dissociation} \small Correlations between mean participant data and all models that make different predictions for water versus honey trials (in both experiments). A and B show the IFE models and ground truth, while C and D show the alternative models (ConvNet, MarbleSim, and SimpleSim). The alternative models exclude the gravity heuristic, as it makes the same predictions for both water and honey. In Experiment 2, the bars with small or negative values (in B) establish that participants' responses are sensitive to the stickiness and viscosity of a liquid. Model 1 and Model 2 refer to the two different versions of IFE honey. Model 1 simulates honey as a fluid with high particle friction and stickiness, while Model 2 simulates honey as water with high damping. Summary of parameters for all models are found in Table~\ref{tab:parameters_summary}.}
\end{figure}

Importantly, our results show a significant correlation (two-sample Kolmogorov-Smirnov \textit{p} $<$ .0001) between individual participants' judgments and the ground truth in all our experiments (see Fig~\ref{fig:indiv_vs_null_hist}). The null hypothesis was generated by drawing random samples from a beta distribution fit to participant data, which captured the variance of their judgments but not any stimulus-specific structure. This result shows that participants cannot simply do well by chance.

Next we asked: How well are participants' judgments explained by the IFE models versus alternative models? Overall, the results show that the IFE did well in explaining the data, but we also found that the honey IFEs (and honey ground truth) did moderately well in accounting for responses on water trials. Thus, from Experiment 1 alone, it is not clear to what extent people were sensitive to the different fluid properties. In addition, our alternative models also performed moderately well in some cases, compared to the IFE. Experiment 2 addresses both of these issues. Experiment 1 scenes were not designed to guarantee that honey and water resulted in very different outcomes, and thus responses for water versus honey in many scenes \textit{should} be similar. In Experiment 2, we design scenes for which water and honey simulations have maximally different outcomes, and therefore, if people's predictions are sensitive to different liquid properties, we should be able to detect this. Furthermore, the Experiment 2 scenes are a more stringent test of the alternative models, whose physics do not as closely match the ground truths for water or honey as the IFE's does.

Correlations between the data in Experiment 1a and uncertain IFE were $r=0.92[0.91, 0.93]$, $r=0.76[0.74, 0.78]$, and $r=0.81[0.79, 0.82]$ for water, Model 1 honey, and Model 2 honey, respectively. Experiment 1b had correlations of $r=0.74[0.71, 0.78]$, $r=0.79[0.76, 0.83]$, and $r=0.86[0.83, 0.88]$ for water, Model 1 honey, and Model 2 honey, respectively. Fig~\ref{fig:alpha_vs_zeta} summarizes how IFE fits vary across different values of $\zeta$ and $\alpha$. 

\begin{figure}[tp]
\centering
\includegraphics[width=0.6\linewidth]{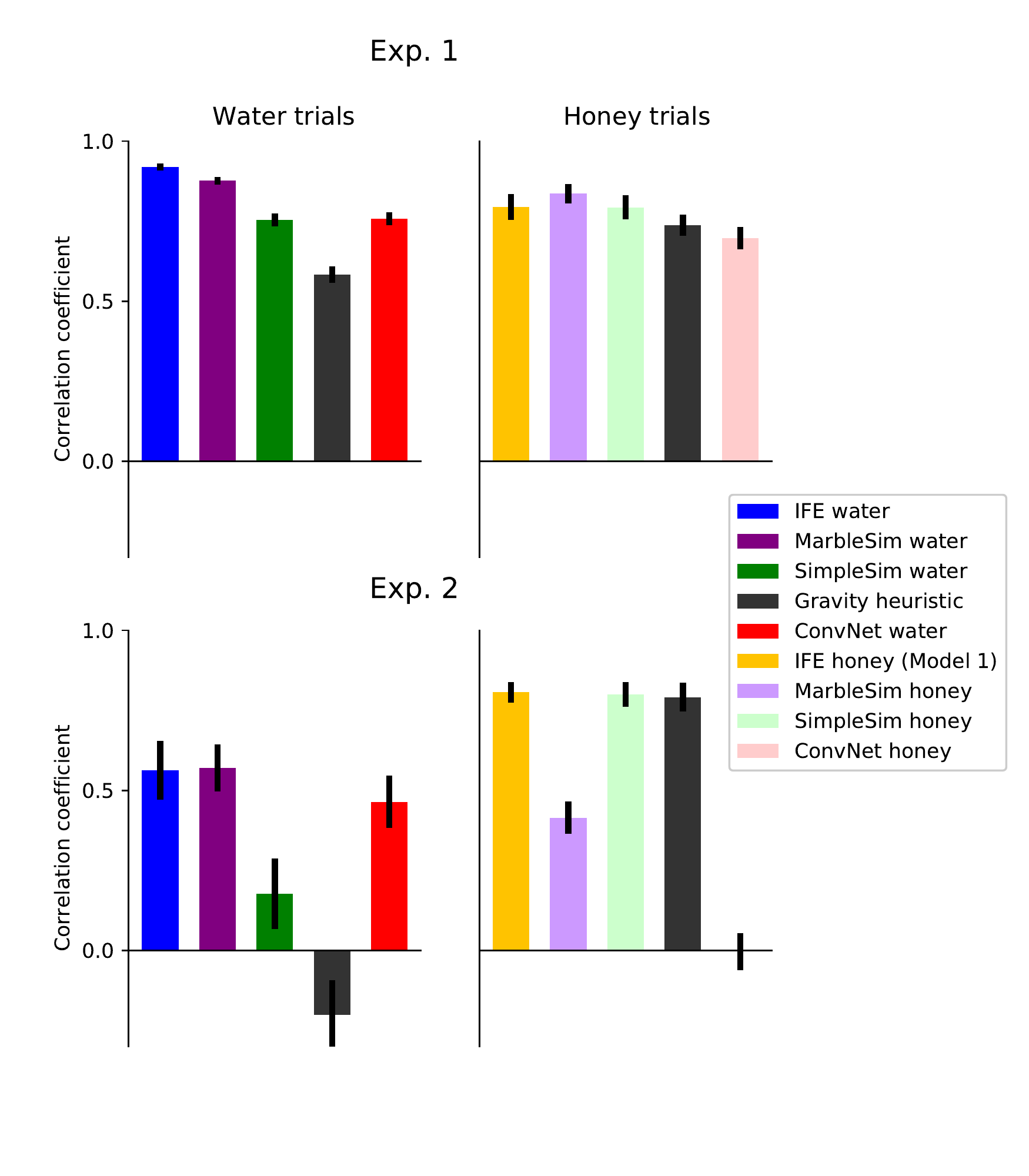}
\caption{\label{fig:IFE_vs_nonsim} \small Comparison of the best-fitting versions of the IFE model and alternative models. All bars for the IFE, MarbleSim, SimpleSim, and ConvNet are identical to their corresponding bars in \protect Fig~\ref{fig:double_dissociation}, and their values are given in Table~\ref{tab:parameters_summary}.}
\end{figure}

\begin{figure}[tp]
\centering
\includegraphics[width=0.9\linewidth]{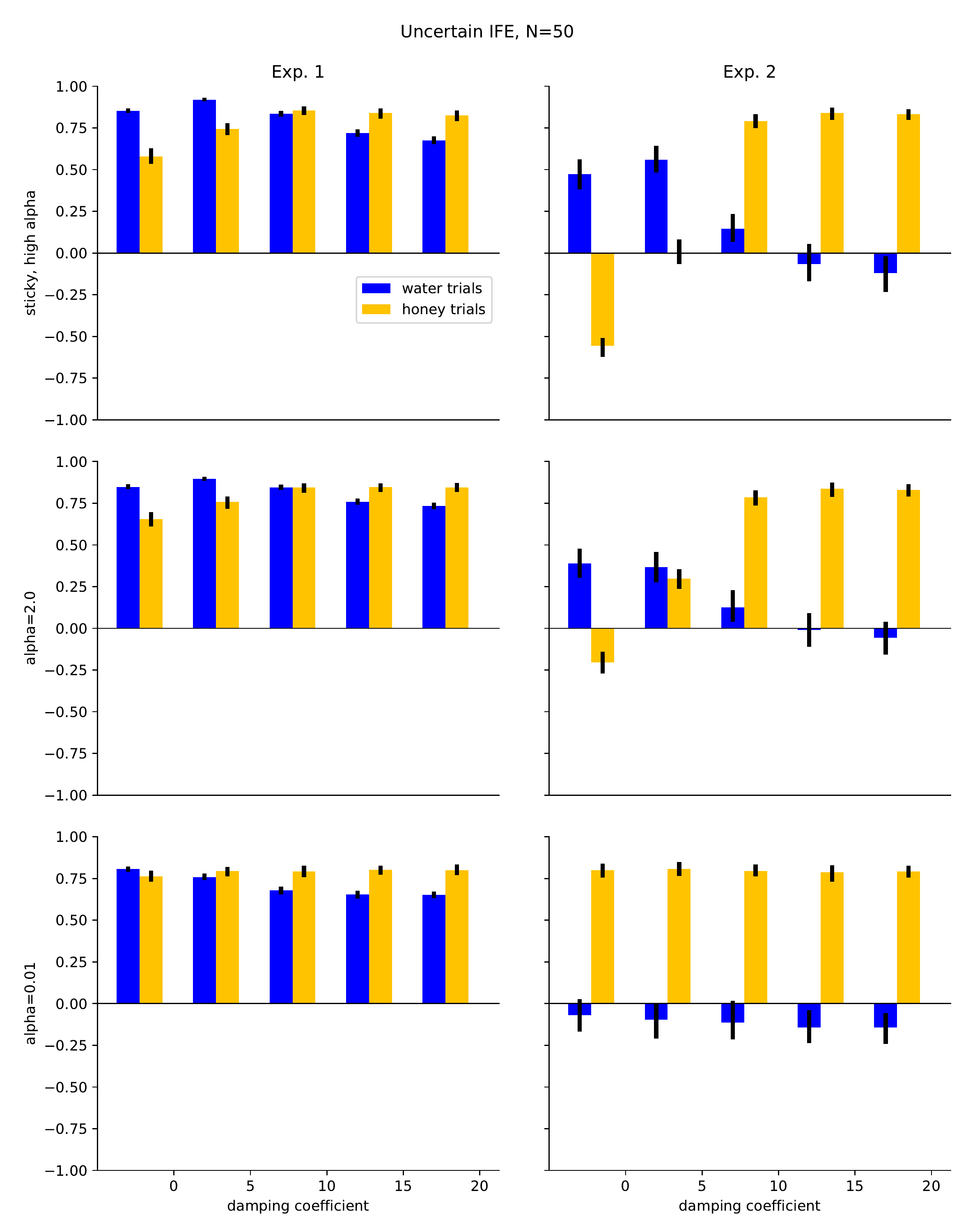}
\caption{\label{fig:alpha_vs_zeta} \small Uncertain IFE performance across a range of $\alpha$ and damping values at 50 particles. Results are qualitatively similar to the deterministic IFE, but with higher correlations.}
\end{figure}

The IFE outperformed all alternative models in Experiment 1a, but was slightly outperformed by MarbleSim in Experiment 1b. MarbleSim was also highly competitive in water trials. SimpleSim was somewhat competitive in water trials and highly competitive in honey trials, and the gravity heuristic was competitive on honey trials, but less competitive in water trials. Table~\ref{tab:altmodels_summary1} summarizes the numerical results of the alternative models, which are also depicted visually in Fig~\ref{fig:double_dissociation}C (with the exception of the gravity heuristic, which is shown in Fig~\ref{fig:IFE_vs_nonsim}). Table~\ref{tab:altmodels_summary1} also summarizes the agreements between alternative models and the ground truth predictions for each liquid. 

MarbleSim's best-fitting parameter values for water and honey trials respectively were ($N=25$, $\zeta=2$) and ($N=75$, $\zeta=6$). Both $N$ and $\zeta$ were fit independently for water versus honey trials. We will refer to these best-fit parameter values as ``water MarbleSim'' and ``honey MarbleSim'', respectively. SimpleSim's best-fitting parameter values for water and honey trials respectively were ($g=0.1$, $m=0.1$) and ($g=0.0$, $m=0.04$). Both $g$ and $m$ were fit independently for water versus honey trials. We will refer to these best-fit parameter values as ``water SimpleSim'' and ``honey SimpleSim'', respectively. To implement SimpleSim's uncertainty over $m$, which was analogous to the IFE's and MarbleSim's uncertainty over $\zeta$, we average model predictions over the range $[m - \sigma_m, m + \sigma_m]$, where $\sigma_m / (m_{max} - m_{min}) = \sigma / (\zeta_{max} - \zeta_{min})$, and max and min represent the maximum and minimum values explored for those parameters. Thus, since $m_{min}=0$, $m_{max}=0.2$, $\zeta_{min}=0$, $\zeta_{max}=20$, and $\sigma=4$, then $\sigma_m=0.04$. Here, we only report results for this value of $\sigma_m$.

The ConvNet was initially trained on the Experiment 1a ground truth (``water ConvNet''), and a separate model (``honey ConvNet'') was then made by fine-tuning the original network on a smaller number of honey ground truth simulations, in order to better capture human judgments on Experiment 1b. As seen in Table~\ref{tab:altmodels_summary1}, the water ConvNet successfully learned to approximate ground truth predictions, but its correlation with people in water trials was comparable to ground truth itself. The honey ConvNet was less successful in learning the ground truth, but its correlations with people in Experiment 1b were also comparable to those with the ground truth. Like with the honey ground truth, the honey ConvNet's predictions were highly correlated with people in Experiment 1a. Experiment 2 further addresses this issue, by establishing whether people actually make different predictions for water and honey.

\begin{table}
\begin{adjustwidth}{-2.25in}{0in}
\begin{center}
\begin{tabular}{lrrrr}
\hline
 \textbf{Exp. 1}   &                     &                     &                &                \\
                   & Data (water trials) & Data (honey trials) & Ground (water) & Ground (honey) \\
 Water ConvNet     & 0.76[0.74, 0.78]    & 0.52[0.46, 0.56]    & 0.91           & 0.69           \\
 Honey ConvNet     & 0.85[0.83, 0.86]    & 0.70[0.65, 0.74]    & 0.76           & 0.82           \\
 Water SimpleSim   & 0.75[0.73, 0.77]    & 0.66[0.62, 0.71]    & 0.56           & 0.60           \\
 Honey SimpleSim   & 0.68[0.66, 0.70]    & 0.76[0.73, 0.80]    & 0.38           & 0.60           \\
 Water MarbleSim   & 0.88[0.86, 0.89]    & 0.76[0.72, 0.79]    & 0.79           & 0.78           \\
 Honey MarbleSim   & 0.83[0.81, 0.84]    & 0.84[0.81, 0.87]    & 0.60           & 0.78           \\
 Gravity heuristic & 0.58[0.55, 0.60]    & 0.73[0.70, 0.77]    & 0.19           & 0.53           \\
\hline
\end{tabular}
\caption{Correlations between alternative models and subject data and the ground truth predictions}
\label{tab:altmodels_summary1}
\end{center}
\end{adjustwidth}
\end{table} 

We also found hints that people's predictions were sensitive to the physical attributes of the liquids, which could be predicted by the IFE. In both Experiment 1a and 1b, participants were at least partially sensitive to the physical attributes of viscosity and stickiness of each liquid: the ground truth model whose particle friction and stickiness corresponded to Experiment 1a was a better fit to Experiment 1a's participants' responses than those of Experiment 1b (see Fig~\ref{fig:double_dissociation}). Experiment 1b's participants' responses were better fit by the ground truth model with Experiment 1b's physical attributes. However, the fit between Experiment 1a responses and the honey ground truth was quite high, an issue which we address in Experiment 2.

In both experiments, including damping in the simulation models improved fits with people's judgments. Furthermore, the uncertain IFE model had modestly higher correlations in most cases than the deterministic IFE (results not presented here). However, future work should further investigate the role of uncertainty in this kind of task. We also tested our previously presented uncertainty model \cite{bateshumans}, which assumed veridical physics plus uncertainty about the initial location of the fluid. We found this model to be more consistent with the data than ground truth, but it did not account as well for the data as either the deterministic or uncertain damped IFE models we present here.

In Experiment 1a, split-half correlations reveal participants were highly consistent with each other ($r=0.96[0.94, 0.97]$). Experiment 1b participants were less consistent with each other ($r=0.81[0.73, 0.86]$) than Experiment 1a, which might be attributable to less familiarity with the liquid, since Experiment 1a participants saw a liquid that behaved similarly to real water, but the liquid in Experiment 1b, while comparable to a highly viscous honey, was less similar to most common liquids. In addition, real honey has highly variable viscosity, so any experience people have with honey is less consistent than with water.

\begin{table}
\begin{adjustwidth}{-2.25in}{0in}
\begin{center}
\begin{tabular}{lrrrrrrr}
\hline
 \textbf{Model}    & \textbf{N} & $\boldsymbol\alpha$ & $\boldsymbol\zeta$ & $\boldsymbol\sigma$ & \textbf{sticky} & $\boldsymbol m$    & $\boldsymbol g$ \\
\hline
 Water IFE         & 50  & 0.01                    & 5       & 4        & No     & N/A    & N/A \\
 Honey 1 IFE       & 50  & 1.25(Exp. 1)/2.0(Exp. 2)& 5       & 4        & Yes    & N/A    & N/A \\
 Honey 2 IFE       & 50  & 1.25(Exp. 1)/2.0(Exp. 2)& 11      & 4        & No     & N/A    & N/A \\
 Water MarbleSim   & 25  & N/A                     & 2       & 4        & No     & N/A    & N/A \\
 Honey MarbleSim   & 75  & N/A                     & 6       & 4        & No     & N/A    & N/A \\
 Ground truth water& 200 & 0.01                    & 0       & 0        & No     & N/A    & N/A \\
 Ground truth honey& 100 & 1.25(Exp. 1)/2.0(Exp. 2)& 0       & 0        & Yes    & N/A    & N/A \\
 Water SimpleSim   & N/A & N/A                     & N/A     & 4        & N/A    & 0.1    & 0.1 \\
 Honey SimpleSim   & N/A & N/A                     & N/A     & 4        & N/A    & 0.04   & 0.0 \\
 Gravity heuristic & N/A & N/A                     & N/A     & N/A      & N/A    & 0.0    & 0.0 \\
 ConvNet           & N/A & N/A                     & N/A     & N/A      & N/A    & N/A    & N/A \\
\hline
\end{tabular}
\caption{Summary of the model parameters used in figures 6 and 7. The same set of parameters was used to model both experiments (with the exception of $\alpha$ in IFE honey models, which was fixed a priori).}
\label{tab:parameters_summary}
\end{center}
\end{adjustwidth}
\end{table} 

\subsection*{Experiment 2}

Our results show that people's judgments on trials of a particular fluid type were positively correlated with the appropriate model, and near zero or anti-correlated with the other fluid type (see Fig~\ref{fig:double_dissociation}B). 

Correlations with the IFE model were calculated using the same model parameters from Experiment 1. The correlations for participants on water trials were $r=0.57[0.48, 0.64]$ with the water IFE, $r=-0.10[-0.20, 0.01]$ with the Model 1 honey IFE and $r=0.08[-0.03, 0.19]$ with Model 2 honey IFE. On honey trials, correlations were $r=0.00[-0.07, 0.08]$ with the water IFE, $r=0.81[0.77, 0.85]$ with the Model 1 honey IFE, and $r=0.82[0.78, 0.86]$ with Model 2 honey IFE. Correlations with ground truth for water trials were $r=0.41[0.35, 0.47]$ and $r=-0.03[-0.08, 0.03]$ for water and honey respectively. Correlations with ground truth for honey trials were $r=-0.69[-0.71, -0.68]$ and $r=0.69[0.68, 0.7]$ for water and Model 1 honey respectively. 
 
As shown in Fig~\ref{fig:IFE_vs_nonsim}, the gravity heuristic and SimpleSim accounted just as well for honey trial data as the honey IFE models, but were not as well correlated with water trials. Numerical results for all alternative models are also summarized in Table~\ref{tab:altmodels_summary2}, and visualized in Fig~\ref{fig:double_dissociation}D. Parameters for MarbleSim and SimpleSim were fixed to their corresponding values in Experiment 1. Results show that, in contrast to Experiment 1, MarbleSim did poorly on honey trials and SimpleSim did poorly on water trials, compared to the IFE. 

As in Experiment 1, the ConvNet model was trained initially on water trials, and then fine-tuned on a smaller number of honey trials, resulting in two separate networks. (Note that separate networks needed to be trained for Experiment 2, because there was a cup rather than a divider at the bottom.) The ConvNet results are shown in Table~\ref{tab:altmodels_summary2}. Like Experiment 1, the water ConvNet was quite successful in predicting ground-truth physics, but its correlation with the water trial responses were still lower than the water IFE. By contrast, the honey ConvNet was not as successful in predicting the ground truth on our stimuli. We believe that this is because the scenes we picked as stimuli were not drawn from the same distribution as those in the ConvNet training set. Specifically, the stimuli were drawn from a skewed distribution, which filtered for scenes with highly divergent outcomes for water versus honey, whereas the training set as a whole was not. We believe this problem did not affect the water ConvNet as much, because of differences in behavior between the water and honey. Honey may exhibit different ``regimes'' of behavior: if only a low proportion of the particles get stuck on an obstacle surface, the simulation may have a very different outcome than if slightly more particles get stuck. It is possible that with a very large training set, the network would be more accurate on our honey stimuli. However, we stopped seeing incremental improvement when the honey ConvNet test set (in both experiments) included more than about 25\% of the full set of 10000 training examples.

\begin{table}
\begin{adjustwidth}{-2.25in}{0in}
\begin{center}
\begin{tabular}{lrrrr}
\hline
 \textbf{Exp. 2}   &                     &                     &                &                \\
                   & Data (water trials) & Data (honey trials) & Ground (water) & Ground (honey) \\
 Water ConvNet     & 0.46[0.36, 0.55]    & -0.56[-0.62, -0.51] & 0.91           & -0.49          \\
 Honey ConvNet     & 0.13[0.03, 0.23]    & -0.00[-0.08, 0.07]  & 0.05           & 0.15           \\
 Water SimpleSim   & 0.22[0.12, 0.32]    & 0.58[0.52, 0.64]    & -0.40          & 0.49           \\
 Honey SimpleSim   & -0.06[-0.16, 0.05]  & 0.78[0.74, 0.82]    & -0.76          & 0.70           \\
 Water MarbleSim   & 0.56[0.48, 0.64]    & -0.16[-0.23, -0.09] & 0.56           & -0.08          \\
 Honey MarbleSim   & 0.33[0.23, 0.42]    & 0.41[0.35, 0.47]    & -0.09          & 0.54           \\
 Gravity heuristic & -0.22[-0.32, -0.12] & 0.79[0.74, 0.83]    & -0.89          & 0.81           \\
\hline
\end{tabular}
\caption{Correlations between alternative models and subject data and the ground truth predictions}
\label{tab:altmodels_summary2}
\end{center}
\end{adjustwidth}
\end{table}

The results of Experiment 2 show that people are sensitive to the physical parameters (both stickiness and viscosity) of each fluid. Furthermore, our data show that people can accommodate their predictions systematically to different fluids with different physical parameters, but they may do so in one of two ways: by adjusting parameters in the simulation corresponding roughly to the physics, like stickiness and viscosity, or by approximating these physical variations using a single simple parameter, damping. The latter is cognitively appealing for its representational simplicity, but needs further investigation as a candidate cognitive model.

But while participants had good fits with both honey IFE models on honey trials, the water IFE model's fit to participants' responses was significantly lower than in the previous experiment, and responses were less accurate with respect to ground truth. In addition, subjects' split-half correlations were much lower ($r=0.72[0.53, 0.82]$, as compared to $r=0.96[0.94, 0.97]$ in Experiment 1 water trials, $r=0.81[0.73, 0.86]$ in Experiment 1 honey trials, and $r=0.87[0.81, 0.92]$ in Experiment 2 honey trials). One potential explanation of this performance gap is that the task was more intrinsically difficult, as the cup was less wide than each bin in Experiment 1. That is, Experiment 2 scenes may require more precision in simulation (i.e. more samples) to match the accuracy levels in Experiment 1. As a measure of how much precision was required, we compared IFE prediction variance across tasks. That is, we looked at the deterministic IFE predictions over a range of $\zeta$ values to see how much they varied. If there is high variance, it means the model is less ``certain'' of the outcome, and therefore the task is more difficult. The ranges of $\zeta$ examined were the same as those used by the uncertain IFE in the results presented above (i.e. $1 \leq \zeta \leq 9$ for water and Model 1 honey, $7 \leq \zeta \leq 15$ for Model 2 honey). Table~\ref{tab:IFE_variance_table} shows the mean variance across scenes for each experiment and IFE fluid type. The mean variance for water in Experiment 2 was significantly higher than the mean variances for the other two IFE models in both experiments (unpaired t-test, all p-values near zero), supporting the notion that Experiment 2 water stimuli were intrinsically more difficult than the other three tasks. (Note that the human performance was higher on Experiment 1 (both water and honey trials) and Experiment 2 honey trials, but lower on Experiment 2 water trials, qualitatively matching the pattern of model variances.) However, further work will be necessary to understand why the model had higher agreement with ground truth in Experiment 2 water trials than human responses. 

\begin{table}[h]
\begin{center}
\begin{tabular}{llll}
\hline
& Water & Honey (Model 1) & Honey (Model 2) \\ 
\hline
Experiment 1 & 0.003[0.003, 0.003] & 0.002[0.002, 0.002] & 0.003[0.003, 0.004] \\
 Experiment 2 & 0.014[0.013, 0.016] & 0.001[0.001, 0.001] & 0.009[0.008, 0.010] \\
\hline
\end{tabular}
\caption{IFE variance across experiments.}
\label{tab:IFE_variance_table}
\end{center}
\end{table}

\section*{Discussion}

Here we introduced the first computational cognitive model of how humans judge fluid dynamics in complex everyday settings. Our proposed model, the ``Intuitive Fluids Engine'' (IFE), holds that humans run particle-based simulations to predict how a fluid will flow in the world. Our experimental results show that humans are able to make coarse, but often accurate, predictions about fluid dynamics, which are best explained by the IFE, compared to several alternatives, including coarser simulation as well as models that do not explicitly simulate liquid dynamics. In two experiments, we found that people were able to predict fluid flow in a range of novel environments with multiple obstacles, varied fluid properties, and different task dynamics. Their predictions demonstrated that they could take the complex geometric structure into account, as well as fluid properties such as stickiness and viscosity, and these phenomena were captured by our IFE model. Solving these prediction tasks accurately would be very computationally intensive even for an agent with perfect knowledge of the ground truth physics, and our model explains the difference between ground truth accuracy and human performance as resulting from limited numbers of particles used to represent the fluid, from a damping term that modulated the momentum of the simulated particles, and from imperfect approximations to the physics inherent to all particle-based simulators. Together people's general competence and their adherence to our model's predictions are consistent with the theory that important aspects of human physical scene understanding can be accounted for by cognitive mechanisms of approximate probabilistic simulation \cite{battaglia2013simulation}, and extends this thesis to the more complex and under-studied domain of how humans understand and interact with fluids.

A crucial feature of our IFE model is how coarsely it approximates the true physics of fluids. Though our particle-based model is in the same family as SPH methods used for approximate simulations in physics, engineering, computer graphics and video games, its details and implementation differ importantly from those used in other disciplines. The rules that govern the particle-particle and particle-rigid interactions, the parameters of the simulated materials, and the granularity of the mass density approximation are highly simplified relative to those used in typical engineering applications, in ways that render the simulation non-veridical even though it still looks reasonably natural. In particular, very small numbers of particles, by the standards of typical SPH methods (e.g., 50 in our model, versus the hundreds of thousands or millions of particles typical in graphics and engineering, or the thousands typical in video games), gave the best fit to people's judgments.  And when simulating viscous fluids, a variant of our model that lumps various ``stickiness'' parameters into a single damping parameter fits human judgments well compared to the variant that keeps them separate. Though it requires further exploration, representational simplifications like these may point toward ways that human minds simulate complex real-world physics with cognitively plausible and efficient computations. 

The only exception to the general pattern of agreement between physical simulation and people's judgments occurred in the "Water" condition of Experiment 2, where a minority of participants had no correlation with ground truth. Our analyses showed that trials in this condition were objectively more difficult for a simulation-based approach than those in other conditions, but it is also possible that there are interesting qualitative differences between these trials to be explored further. Future work should more systematically investigate people's patterns of success and failures in predicting qualitatively different kinds of fluid motion, and we expect that  this will yield more insights into the nature of the computations and representations that people use in predicting fluid flow. 

A common critique of quantitative simulation-based cognitive models of intuitive physics \cite{forbus1997qualitative,davis2015scope} is they are impractical or unrealistic, due to the computation costs and the detailed physical knowledge implicitly embedded in these simulation algorithms.  The simulation-based model we propose here should alleviate some of these concerns: Even extremely simplified simulations, with very coarse physical approximations, can nonetheless capture the fluid dynamics well enough for everyday judgment purposes.  At the same time, these models can capture to a reasonable extent the surprisingly accurate quantitative predictions of fluid flow that people are able to make.  

The alternative models we presented aimed to provide a benchmark, against which to compare the performance of our IFE results. We explored simpler, particle-based simulation models, and found that they captured certain aspects of people's judgments, but no single model could account as fully for people's predictions as the IFE. Nonetheless, simpler simulation models can do reasonably well here, which suggests that the brain may implement something not quite as sophisticated as our IFE. Taken together, the success of our IFE and alternative simulation models provides strong evidence for relatively sophisticated mental simulation in people's heads, but point toward the need to further explore simpler simulation alternatives that are still sophisticated enough to account for the wide range of accurate predictions that people can make.

Our SimpleSim model performed especially well on honey trials in both experiments, achieving correlations similar to those of the IFE. A possible explanation for this success is that honey is an easier fluid to simulate heuristically. Supporting this notion, SimpleSim in Experiment 2 had a maximum correlation with the honey ground truth of $r=0.80$ (occurring at $g=0.1$, $m=0.06$), but with the water ground truth, the maximum correlation was just $r=0.20$ (occurring at $g=0.0$, $m=0.2$). (Recall that Experiment 2 scenes were specifically designed to elicit different outcomes for water versus honey, and are thus the best test of whether any parameter setting of SimpleSim could approximate water, as compared to honey.) This result makes intuitive sense, since both real honey and particles in SimpleSim tend to fall straight down near obstacle edges, by contrast to water, which splashes energetically, in different and less predictable directions. We believe that the moderate success of the heuristic simulation model can be attributed in part to the above: it is intrinsically well-suited for simulating sticky, viscous liquids falling under gravity. SimpleSim's relative success in Experiment 1 water trials can largely be attributed to its close agreement with honey ground truth physics, since honey ground truth and honey IFE also had high correlations with those trials. Thus, the real test of whether SimpleSim is a good model for water trials is Experiment 2, where it does poorly relative to water IFE, water ground truth, and water MarbleSim.

We also explored a non-simulation, neural network model that learned a fast mapping between images of the starting configurations and the numerical outcomes. First, our neural network model shows that it is possible for a connectionist architecture to learn an accurate mapping from a still image representing initial conditions to a single number representing the simulated outcome for our task. Since our networks were trained on ground truth physics, they learned close approximations to the ground truth. It is perhaps unsurprising, then, that they had similar correlations with human data compared to the ground truth simulations. If people learn a similar kind of fast mapping between starting conditions and final outcomes, our results suggest there may be other biases or constraints involved, which we have not captured here. (We also tested whether any intermediate points in training, prior to convergence, resulted in higher correlations with subjects, and found this not to be the case).

Previous AI models for explicit reasoning about fluids (e.g. \cite{kim1993qualitative,kim1990qualitative, collins1987reasoning, davis2008liquids}, \cite{gardin1989analogical}) have been studied in fundamentally different contexts than those we examined here. For example, \cite{kim1990qualitative} and \cite{kim1993qualitative} assume laminar flow that is highly constrained by container geometry (e.g. a piston cylinder). These models were designed to answer binary questions (Is the fluid flowing up or down? Is pressure increasing or decreasing?), rather than to predict the location of liquid during or after a splash. It is not clear how such qualitative or logical modeling approaches could be adapted to the present experiment, which involves fluids in rapid dynamical flow, or whether they could capture our participants' quantitative judgments, but this is an important topic for future work. Perhaps the most similar previous model is \cite{gardin1989analogical}, which could be seen as a simpler particle-based approach and in part motivated the gravity heuristic and SimpleSim models here. Their approach may be adaptable to the present experiments, but it is not immediately clear how to modify it such that it could distinguish between different liquids, such as water and honey.

Future work should explore how the precision, temporal duration, and other structural and parametric features of mental simulation for fluids might be implemented. For example, how does the precision vary as a function of how far into the future one must mentally simulate? \cite{kubricht2016probabilistic} showed that for stability judgments, where the question is how far a fluid-containing vessel can be tilted before the fluid spills out and only one-time step predictions are needed (in contrast to our judgments here which required looking ahead many time-steps), IFE models can be extremely quantitatively accurate as accounts of people's judgments. How closely do the attributes represented by the mind correspond to the actual physical characteristics of a liquid? Physically accurate simulation models can account for people's judgments, but we also showed that a simpler simulation model (SimpleSim) also did reasonably well, even if the more physically accurate simulation models did better overall. This raises the question as to whether other simple simulation models may do better than SPH in accounting for people's judgments.

Can our general approach extend to capture other classes of physical intuitions that go beyond rigid-body dynamics? Particle-based models can provide reasonable simulations for liquids and gases, as well as collections of solid elements (e.g., piles of sand) and composite materials (e.g., mashed potatoes or Play-Doh) whose dynamics share similarities with liquids. In fact, work by \cite{kubricht2017consistent} provides preliminary evidence that particle-based simulation models can capture people's judgments about sand in a very similar task to the one we present here. What are the limits of simulation-based models--what kinds of non-rigid dynamics can people make coherent predictions about, and might some be better explained by alternative approaches such as qualitative reasoning? Do people begin to use some form of qualitative reasoning as simulation becomes too computationally demanding or in situations where they have insufficient prior experience? There is evidence that internal forward models in the motor system are involved in predicting dynamics that cannot be reenacted by the body \cite{schubotz2007prediction}, which might help explain how mental simulations are implemented. Establishing connections between such work and our approximate simulation framework could lead to a deeper understanding of people's capabilities and limitations when simulating physics.

Another important question is: How could intuitive physics engines be represented in the brain? \cite{fischer2016functional} report fMRI evidence for physics-specific processing in an array of cortical regions related to the "multiple demand" system. But no models yet have connected the computational models to their neural substrates. Inspired by the recent successes of deep learning approaches in AI, a plausible candidate might be a recurrent neural network with dynamic, parallel, and distributed structure~\cite{michalski2014modeling}. To date, neural network models have only been able to capture simple and restricted classes of rigid-body dynamics~\cite{grzeszczuk1998neuroanimator,lerer2016learning}, but as work on multimodal image annotation, linguistic modeling, and more general purpose distributed algorithm learning systems (e.g., LSTMs~\cite{hochreiter1997long,graves2012supervised}, the ``Neural Turing Machine'' \cite{graves2014neural}) advances, more richly structured scenes and physical dynamics might become accessible to these frameworks. A more general question for development, and computational cognitive psychologists is: Where does people's knowledge of liquids come from? Five month-olds can distinguish between solids and liquids in novel contexts after observing their distinct patterns of movement \cite{hespos2009five,hespos2015five}, which suggests either a very data-efficient experience-based learning process or innate biases. Perhaps our simulation-based model's core components can be measured in young children, or its more complex features can be observed as they emerge and mature.

\section*{Conclusion}

Approximate simulation is a powerful framework that may explain how people understand a wide variety of complex physical processes they encounter in everyday life, such as rigids, non-rigids, and fluids. This work offers the first computational model of how people make intuitive predictions about fluids, and provides evidence that a particle-based simulation model can account for human predictions about liquids with different substance attributes.

\section*{Supporting information}





\paragraph*{S1 Appendix.}
\label{S1_Appendix}
{\bf Simulation in Blender}
The stimuli in both experiments used Blender's (www.blender.org) lattice-Boltzman liquid simulator. All simulations used a grid resolution of 300 and 100 tracer particles. The water stimuli used Blender's preset kinematic viscosity of $1.000$ centiPoise. The honey stimuli had a kinematic viscosity of $100.0$ cP in Experiment 1 and $2.000$ cP in Experiment 2. All other simulation settings were left to their default values (Real World Size: 0.5 meters; Grid Levels: -1; Compressibility: 0.005; Slip Type: Free Slip; Surface Smoothness: 1.000; Surface Subdivisions: 0). Despite being set to free slip, the simulations at high viscosity liquid exhibited adhesive effects. The liquids were rendered using a Mix Shader with 0.05 fraction Diffuse BSDF (Roughness: 0.000) and Beckmann Glass BSDF (Roughness: 0.061, IOR: 1.330).



\nolinenumbers

%
%
%

\bibliography{refs}

\begin{thebibliography}{10}

\bibitem{sanborn2013reconciling}
Sanborn AN, Mansinghka VK, Griffiths TL.
\newblock Reconciling intuitive physics and Newtonian mechanics for colliding
  objects.
\newblock Psychological review. 2013;120(2):411.

\bibitem{sanborn2014testing}
Sanborn AN.
\newblock Testing Bayesian and heuristic predictions of mass judgments of
  colliding objects.
\newblock Frontiers in psychology. 2014;5.

\bibitem{battaglia2013simulation}
Battaglia P, Hamrick JB, Tenenbaum JB.
\newblock Simulation as an engine of physical scene understanding.
\newblock Proceedings of the National Academy of Sciences.
  2013;110(45):18327--18332.

\bibitem{gerstenberg2012noisy}
Gerstenberg T, Goodman N, Lagnado D, Tenenbaum J.
\newblock Noisy Newtons: Unifying process and dependency accounts of causal
  attribution.
\newblock In: Proceedings of the Annual Meeting of the Cognitive Science
  Society. vol.~34; 2012.

\bibitem{smith2013consistent}
Smith K, Battaglia P, Vul E.
\newblock Consistent physics underlying ballistic motion prediction.
\newblock In: Proceedings of the 35th Conference of the Cognitive Science
  Society; 2013. p. 3426--3431.

\bibitem{smith2013sources}
Smith KA, Vul E.
\newblock Sources of uncertainty in intuitive physics.
\newblock Topics in cognitive science. 2013;5(1):185--199.

\bibitem{hamrick2016inferring}
Hamrick JB, Battaglia PW, Griffiths TL, Tenenbaum JB.
\newblock Inferring mass in complex scenes by mental simulation.
\newblock Cognition. 2016;157:61--76.

\bibitem{bender2011sph}
Bender J, Erleben K, Galin E.
\newblock SPH based shallow water simulation.
\newblock In: Workshop on Virtual Reality Interaction and Physical Simulation;
  2011.

\bibitem{akinci2013versatile}
Akinci N, Akinci G, Teschner M.
\newblock Versatile surface tension and adhesion for SPH fluids.
\newblock ACM Transactions on Graphics (TOG). 2013;32(6):182.

\bibitem{mccloskey1983naive}
McCloskey M, Kohl D.
\newblock Naive physics: the curvilinear impetus principle and its role in
  interactions with moving objects.
\newblock Journal of Experimental Psychology: Learning, Memory, and Cognition.
  1983;9(1):146.

\bibitem{mccloskey1983intuitive}
McCloskey M, Washburn A, Felch L.
\newblock Intuitive physics: The straight-down belief and its origin.
\newblock Journal of Experimental Psychology: Learning, Memory, and Cognition.
  1983;9(4):636.

\bibitem{kaiser1986intuitive}
Kaiser MK, Jonides J, Alexander J.
\newblock Intuitive reasoning about abstract and familiar physics problems.
\newblock Memory \& Cognition. 1986;14(4):308--312.

\bibitem{cook1994constructing}
Cook NJ, Breedin SD.
\newblock Constructing naive theories of motion on the fly.
\newblock Memory \& Cognition. 1994;22(4):474--493.

\bibitem{kaiser1992influence}
Kaiser MK, Proffitt DR, Whelan SM, Hecht H.
\newblock Influence of animation on dynamical judgments.
\newblock Journal of experimental Psychology: Human Perception and performance.
  1992;18(3):669.

\bibitem{caramazza1981naive}
Caramazza A, McCloskey M, Green B.
\newblock Naive beliefs in “sophisticated” subjects: Misconceptions about
  trajectories of objects.
\newblock Cognition. 1981;9(2):117--123.

\bibitem{zago2005cognitive}
Zago M, Lacquaniti F.
\newblock Cognitive, perceptual and action-oriented representations of falling
  objects.
\newblock Neuropsychologia. 2005;43(2):178--188.

\bibitem{forbus2011qualitative}
Forbus KD.
\newblock Qualitative modeling.
\newblock Wiley Interdisciplinary Reviews: Cognitive Science.
  2011;2(4):374--391.

\bibitem{davis2008liquids}
Davis E.
\newblock Pouring Liquids: A Study in Commonsense Physical Reasoning.
\newblock Artificial Intelligence. 2008;172:1540--1578.

\bibitem{kim1990qualitative}
Kim H.
\newblock Qualitative Reasoning about Geometry of Fluid Flow.
\newblock Proceedings of Cognitive Science Society-90. 1990; p. 117--124.

\bibitem{kim1993qualitative}
Kim H.
\newblock Qualitative reasoning about fluids and mechanics.
\newblock DTIC Document; 1993.

\bibitem{collins1987reasoning}
Collins JW, Forbus KD.
\newblock Reasoning about Fluids via Molecular Collections.
\newblock In: AAAI; 1987. p. 590--594.

\bibitem{hayes1978naive}
Hayes PJ, et~al.
\newblock The naive physics manifesto.
\newblock Institut pour les {\'e}tudes s{\'e}mantiques et
  cognitives/Universit{\'e} de Gen{\`e}ve; 1978.

\bibitem{gardin1989analogical}
Gardin F, Meltzer B.
\newblock Analogical representations of naive physics.
\newblock Artificial Intelligence. 1989;38(2):139--159.

\bibitem{lerer2016learning}
Lerer A, Gross S, Fergus R.
\newblock Learning Physical Intuition of Block Towers by Example.
\newblock arXiv preprint arXiv:160301312. 2016;.

\bibitem{forbus1997qualitative}
Forbus KD, Gentner D.
\newblock Qualitative mental models: Simulations or memories.
\newblock In: Proceedings of the eleventh international workshop on qualitative
  reasoning. Cortona, Italy; 1997. p. 3--6.

\bibitem{lecun2015deep}
LeCun Y, Bengio Y, Hinton G.
\newblock Deep learning.
\newblock Nature. 2015;521(7553):436--444.

\bibitem{yamins2014performance}
Yamins DL, Hong H, Cadieu CF, Solomon EA, Seibert D, DiCarlo JJ.
\newblock Performance-optimized hierarchical models predict neural responses in
  higher visual cortex.
\newblock Proceedings of the National Academy of Sciences.
  2014;111(23):8619--8624.

\bibitem{battaglia2016interaction}
Battaglia P, Pascanu R, Lai M, Rezende DJ, et~al.
\newblock Interaction networks for learning about objects, relations and
  physics.
\newblock In: Advances in Neural Information Processing Systems; 2016. p.
  4502--4510.

\bibitem{chang2016compositional}
Chang MB, Ullman T, Torralba A, Tenenbaum JB.
\newblock A compositional object-based approach to learning physical dynamics.
\newblock arXiv preprint arXiv:161200341. 2016;.

\bibitem{kawabe2015seeing}
Kawabe T, Maruya K, Fleming RW, Nishida S.
\newblock Seeing liquids from visual motion.
\newblock Vision research. 2015;109:125--138.

\bibitem{paulun2015seeing}
Paulun VC, Kawabe T, Nishida S, Fleming RW.
\newblock Seeing liquids from static snapshots.
\newblock Vision research. 2015;115:163--174.

\bibitem{macklin2014unified}
Macklin M, M{\"u}ller M, Chentanez N, Kim TY.
\newblock Unified particle physics for real-time applications.
\newblock ACM Transactions on Graphics (TOG). 2014;33(4):153.

\bibitem{monaghan2005smoothed}
Monaghan JJ.
\newblock Smoothed particle hydrodynamics.
\newblock Reports on progress in physics. 2005;68(8):1703.

\bibitem{bateshumans}
Bates CJ, Yildirim I, Tenenbaum JB, Battaglia PW.
\newblock Humans predict liquid dynamics using probabilistic simulation.
\newblock Proceedings of the 37th Annual Conference of the Cognitive Science
  Society. 2015; p. 171--177.

\bibitem{monaghan1992smoothed}
Monaghan JJ.
\newblock Smoothed particle hydrodynamics.
\newblock Annual review of astronomy and astrophysics. 1992;30:543--574.

\bibitem{krizhevsky2012imagenet}
Krizhevsky A, Sutskever I, Hinton GE.
\newblock Imagenet classification with deep convolutional neural networks.
\newblock In: Advances in neural information processing systems; 2012. p.
  1097--1105.

\bibitem{jia2014caffe}
Jia Y, Shelhamer E, Donahue J, Karayev S, Long J, Girshick R, et~al.
\newblock Caffe: Convolutional Architecture for Fast Feature Embedding.
\newblock arXiv preprint arXiv:14085093. 2014;.

\bibitem{gigerenzer2011heuristics}
Gigerenzer G, Hertwig R, Pachur T.
\newblock Heuristics: The foundations of adaptive behavior.
\newblock Oxford University Press, Inc.; 2011.

\bibitem{mcclelland2013integrating}
McClelland JL.
\newblock Integrating probabilistic models of perception and interactive neural
  networks: a historical and tutorial review.
\newblock Frontiers in psychology. 2013;4.

\bibitem{efron1994introduction}
Efron B, Tibshirani RJ.
\newblock An introduction to the bootstrap.
\newblock CRC press; 1994.

\bibitem{davis2015scope}
Davis E, Marcus G.
\newblock The Scope and Limits of Simulation in Cognitive Models.
\newblock arXiv preprint arXiv:150604956. 2015;.

\bibitem{kubricht2016probabilistic}
Kubricht J, Jiang C, Zhu Y, Zhu SC, Terzopoulos D, Lu H.
\newblock Probabilistic simulation predicts human performance on viscous
  fluid-pouring problem.
\newblock In: Proceedings of the 38th Annual Conference of the Cognitive
  Science Society; 2016.

\bibitem{kubricht2017consistent}
Kubricht J, Zhu Y, Jiang C, Terzopoulos D, Zhu SC, Lu H.
\newblock Consistent Probabilistic Simulation Underlying Human Judgment in
  Substance Dynamics; 2017.

\bibitem{schubotz2007prediction}
Schubotz RI.
\newblock Prediction of external events with our motor system: towards a new
  framework.
\newblock Trends in cognitive sciences. 2007;11(5):211--218.

\bibitem{fischer2016functional}
Fischer J, Mikhael JG, Tenenbaum JB, Kanwisher N.
\newblock Functional neuroanatomy of intuitive physical inference.
\newblock Proceedings of the National Academy of Sciences. 2016; p. 201610344.

\bibitem{michalski2014modeling}
Michalski V, Memisevic R, Konda K.
\newblock Modeling Deep Temporal Dependencies with Recurrent Grammar Cells.
\newblock In: Advances in neural information processing systems; 2014. p.
  1925--1933.

\bibitem{grzeszczuk1998neuroanimator}
Grzeszczuk R, Terzopoulos D, Hinton G.
\newblock Neuroanimator: Fast neural network emulation and control of
  physics-based models.
\newblock In: Proceedings of the 25th annual conference on Computer graphics
  and interactive techniques. ACM; 1998. p. 9--20.

\bibitem{hochreiter1997long}
Hochreiter S, Schmidhuber J.
\newblock Long short-term memory.
\newblock Neural computation. 1997;9(8):1735--1780.

\bibitem{graves2012supervised}
Graves A, et~al.
\newblock Supervised sequence labelling with recurrent neural networks. vol.
  385.
\newblock Springer; 2012.

\bibitem{graves2014neural}
Graves A, Wayne G, Danihelka I.
\newblock Neural Turing Machines.
\newblock arXiv preprint arXiv:14105401. 2014;.

\bibitem{hespos2009five}
Hespos SJ, Ferry AL, Rips LJ.
\newblock Five-month-old infants have different expectations for solids and
  liquids.
\newblock Psychological Science. 2009;20(5):603--611.

\bibitem{hespos2015five}
Hespos SJ, Ferry AL, Anderson EM, Hollenbeck EN, Rips LJ.
\newblock Five-Month-Old Infants Have General Knowledge of How Nonsolid
  Substances Behave and Interact.
\newblock Psychological Science. 2015;1:13.

\end{thebibliography}






\end{document}